\documentclass[journal]{IEEEtran}

\usepackage{amsmath, amsfonts}            % For advanced math symbols and fonts
\usepackage{array}                        % For advanced table management
\usepackage{graphicx}                     % For including images
\usepackage{url}                          % For URL handling
\usepackage{stfloats}                     % For placement of floats at the bottom
\usepackage{upgreek}                      % For upright Greek letters
\usepackage{textcomp}                     % For text symbols (e.g., text degree)
\usepackage{verbatim}                     % For raw text (e.g., verbatim code)
\usepackage{soul}                         % For highlighting text with \hl
\usepackage[normalem]{ulem}               % For strikethrough and underlining
\usepackage{xfrac}                        % For simple fractions in text
\usepackage{booktabs}                     % For better table styling
\usepackage[ruled]{algorithm2e}           % For algorithms (with ruled lines)
\usepackage{gensymb}                      % For degree symbol and other special symbols
\usepackage{color, colortbl}              % For color and tables
\usepackage{tikz}                         % For drawing figures and custom tables
\usepackage{multirow}                     % For multirow cells in tables
\usepackage{rotating}                     % For rotating elements (e.g., tables)
\usepackage{tabularx}                     % For tables with equal-width columns
\usepackage{tensor}                       % For tensor notation
\usepackage[]{todonotes}                  % For TODO notes
\usepackage{pifont}
\usepackage[caption=false,font=normalsize,labelfont=sf,textfont=sf]{subfig}

\definecolor{mygreen}{rgb}{0.13, 0.75, 0.13}  % Custom green color for checkmarks

\newcommand*{\defeq}{\stackrel{\text{def}}{=}}   % Definition symbol
\newcommand{\tf}[2]{\tensor*[ ^{#2}_{#1} ]{\boldsymbol{T}}{}} % Tensor notation
\newcommand{\tw}{ \mathbf{t} }                   % Translation vector symbol
\newcommand{\refframe}[1]{ \{#1\} }              % Reference frame notation

\newcommand{\invarmoving}{ \boldsymbol{i} }      % Invariant for moving frame
\newcommand{\invarobj}{ \boldsymbol{r}_1 }       % Invariant for object frame

\newcommand{\skewsym}[1]{   \left[{#1}\times\right]}  % Skew symmetric operation
\newcommand{\skewskewsym}[1]{  \left[ {#1}\!\times\!\!\times \right]} % Double skew

\newcommand{\cmark}{\textcolor{mygreen}{\ding{51}}}  % Checkmark symbol
\newcommand{\xmark}{\textcolor{red}{\ding{55}}}      % Cross symbol

\begin{document}
	
	\title{BILTS: A Bi-Invariant Similarity Measure for Robust Object Trajectory Recognition under Reference Frame Variations}
	
	\author{Arno~Verduyn$^{1,2,\dagger}$, Erwin Aertbeli\"en{$^{1,2,\dagger}$}, Glenn~Maes$ ^{1}$, Joris~De~Schutter$^{1,2}$ and Maxim~Vochten$^{1,2}$\\  % <-this % stops a space
	\thanks{This result is part of a project that has received funding from the European Research Council (ERC) under the European Union's Horizon 2020 research and innovation programme (Grant agreement No. 788298).}%
	\thanks{$^1$Department of Mechanical Engineering, KU Leuven, Leuven, Belgium. }%
	\thanks{$^2$Flanders Make at KU Leuven, 3001 Leuven, Belgium.}%
	\thanks{$^\dagger$These authors contributed equally to this work.}%	
	\thanks{email corresponding author: \texttt{arno.verduyn@kuleuven.be}}
	}
	
	\maketitle
	
	\begin{abstract}
	When similar object motions are performed in diverse contexts but are meant to be recognized under a single classification, these contextual variations act as disturbances that negatively affect accurate motion recognition. In this paper, we focus on contextual variations caused by reference frame variations.
	To robustly deal with these variations, similarity measures have been introduced that compare object motion trajectories in a context-invariant manner.
	% invariant trajectory-shape similarity measures hav+e already been proposed.
	However, most are highly sensitive to noise near singularities, where the measure is not uniquely defined, and lack bi-invariance (invariance to both world and body frame variations).
	To address these issues, we propose the novel \textit{Bi-Invariant Local Trajectory-Shape Similarity} (BILTS) measure. Compared to other measures, the BILTS measure uniquely offers bi-invariance, boundedness, and third-order shape identity. Aimed at practical implementations, we devised a discretized and regularized version of the BILTS measure which shows exceptional robustness to singularities. This is demonstrated through rigorous recognition experiments using multiple datasets. On average, BILTS attained the highest recognition ratio and least sensitivity to contextual variations compared to other invariant object motion similarity measures. 
    We believe that the BILTS measure is a valuable tool for recognizing motions performed in diverse contexts and has potential in other applications, including the recognition, segmentation, and adaptation of both motion and force trajectories.
	\end{abstract}
	
	\begin{IEEEkeywords}
	Frame Invariance, Coordinate Invariance, Rigid-body Trajectory, Motion Similarity, Recognition
	\end{IEEEkeywords}
   	
	\section{Introduction}
	\label{sec:introduction}
	%problem and why interesting
	%related work
	%novel thing
	
	%background and motivation
	%problem
	%objective
	%challenges
	%approach
	
	In intelligent robotics and human-robot interaction, there is a need to measure the similarity between new motions and previously established motion models to facilitate motion analysis~\cite{vochten2023invariant}, segmentation~\cite{verduyn2023enhancingmotiontrajectorysegmentation}, and recognition~\cite{Joris2012,DeSchutter2011,Vochten2015,RRV2018,DSRF2018,Lee2018}, as well as motion adaptation~\cite{vochten2019generalizing}. This paper focuses on measuring the similarity between \textit{rigid-body motion trajectories}. Such trajectories are of practical interest since they include the trajectories of manipulated objects, end effectors attached to robot manipulators, and human body parts. 
	% We represent a rigid-body motion by its spatio-temporal trajectory, resulting in a rich and complete representation of the motion.
	
	In dynamic and diverse environments, measuring the similarity between motion trajectories is challenging due to the impact of \textit{diverse contexts}.  These diverse contexts include variations in the \textit{world} frames in which the trajectory coordinates are expressed, which arise due to different camera (or other external sensor) placements and viewpoints. It also includes variations in the \textit{body} frames to express the relative orientation and location of the moving body, which for example arise due to inaccurate body-frame calibrations, inconsistent marker or tracker placements, and partial object occlusions in motion-capture and vision-based systems. Lastly, diverse contexts also include different time profiles of the performed motions. 
	
	When similar motions are performed in diverse contexts but are meant to be recognized under a single classification, the contextual variations act as disturbances that negatively affect accurate motion recognition.

	\subsection{Addressing contextual variations}
	
	To address the challenge of dealing with diverse contexts, methods such as \textit{context alignment}, \textit{machine learning}, and \textit{invariant descriptors} have been proposed.
	
	\textit{Context alignment approaches} aim to align the contexts of the trajectories before measuring similarity. Reference frames can be spatially aligned using affine transformations \cite{wang2008manifold,kabsch1976solution,Rao2002,DSRF2018}. Time profiles can be aligned using algorithms like Dynamic Time Warping (DTW) \cite{sakoe1978dynamic} and Fr\'echet Distance~\cite{TrajectorySimilarityMeasures2021,TrajectoryDistanceSurvey2020}. However, aligning reference frames and time profiles simultaneously remains computationally expensive~\cite{FlexibleSyntacticMatching1999}, limiting this approach to offline applications.
	
	\textit{Machine learning approaches}~\cite{MLreview,ViewInvariantANN2012} aim to identify salient patterns in trajectories that recur across diverse contexts. In these methods, trajectories are considered similar if they share the same salient patterns.
	However, a major challenge lies in preventing these approaches from erroneously identifying patterns within irrelevant contextual features, which can happen when the dataset for training is sparse or biased toward specific contexts~\cite{BiasMitigation}. A potential solution to this challenge involves constructing a dataset that represents a wide range of contexts in an unbiased manner, but this approach can be labor-intensive and may still fall short of capturing all possible contextual variations. Another solution is to implement methods that reduce or completely remove the influence of the context during training.
	
	%\textit{Machine learning approaches} \cite{MLreview,ViewInvariantANN2012} aim to learn general models from example trajectories. They focus on identifying salient features while ignoring variable contextual features. In these approaches, trajectories are deemed similar if they share the same salient features. When provided with a rich dataset, deep learning methods --particularly those using multi-layer networks-- can potentially learn models that interpolate well across a broad range of represented contexts. However, they face the risk of overfitting to the training data. 
	%Furthermore, these models often fail to extrapolate to unrepresented contexts when only a sparse dataset is available for training. 
	
\begin{figure}[t]
		\centering
		\includegraphics[width=\linewidth]{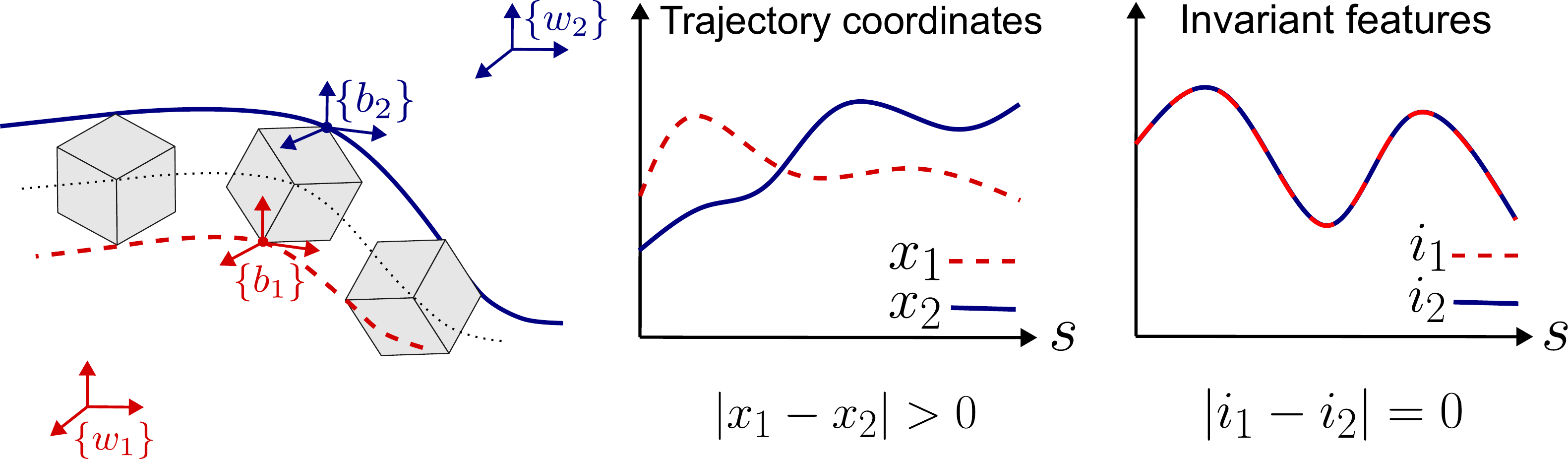}
		\caption{Conceptual figure to visualize the invariant descriptor approach. For a given rigid-body motion, trajectory coordinates (of which only one is visualized above) can vary substantially depending on the selected world reference frame $\{w_1\}$ or $\{w_2\}$ and body reference frame $\{b_1\}$ or $\{b_2\}$. This makes motion similarity measurement based on trajectory coordinates highly frame dependent. To mitigate this dependency, we propose to extract invariant trajectory features which remain unchanged under reference frame transformations. Comparing these invariant features yields a more robust foundation for motion similarity measurement across varying reference frames. }
		\label{fig:introduction}
	\end{figure}

	We focus on \textit{invariant descriptor approaches}~\cite{Lee2018, DeSchutterJoris2010,Vochten2015,vochten2023invariant} which aim to remove the context from the trajectory data to obtain context-invariant trajectory descriptors. % Trajectory descriptors that are invariant to reference frame changes can be obtained by describing the underlying local shape of the trajectory. 
	The similarity between two trajectories can then be defined in an invariant manner by comparing their invariant descriptors. This approach is visualized in Figure~\ref{fig:introduction}. Similarity measures derived from invariant descriptors have shown superior extrapolation capabilities towards unrepresented contexts~\cite{Vochten2015}. In Sec~\ref{sec:overview}, we provide an overview of invariant descriptor approaches.
	
	\subsection{Related work on invariant trajectory descriptors} \label{sec:overview}
	Invariant descriptors for \textit{point trajectories} are based on separating the spatio-temporal trajectory into a spatial curve and a time profile along this curve. Fundamentals of the description of spatial curves were established in the 19\textsuperscript{th} century by Frenet \cite{Frenet1852} and Serret~\cite{Serret1853}, leading to the well-known \textit{Frenet-Serret (FS) formulas}. 
	These formulas show that a world-frame-invariant description of a curve can be found by defining a pair of functions, i.e., the \textit{curvature} and \textit{torsion}, along the curve. These functions characterize the \textit{local shape} of the curve and are obtained by defining a moving frame, referred to as the \textit{FS frame}, along the curve. The curvature and torsion are then obtained by differentiating this frame. 
	
	Based on the FS frame, Cartan \cite{Cartan1937} introduced \textit{the method of the moving frame} in differential geometry for the local examination of submanifolds of various homogeneous spaces. The aim of this method is to provide the submanifold and all its geometrical objects with the most general moving frame. 
	By differentiating this frame, \textit{differential invariants} are obtained that characterize the submanifold apart from transformations embedding it in the surrounding homogeneous space. 
	
	Bishop \cite{bishop1975} defined another moving frame along a spatial curve as a more robust alternative to the Frenet-Serret frame. The resulting frame is referred to as the \textit{Bishop frame} or \textit{rotation-minimizing frame}, since it minimizes the rotation of the frame along the curve. However, it is not a true \textit{local} frame since it depends on the curve's history. %(i.e. initial conditions).
	
	The FS invariants are best known to describe the shape of a point trajectory in Euclidean space. Nevertheless, similar invariants can be derived for the description of the orientation trajectory of a rotating body~\cite{Vochten2015}. The \textit{extended FS (eFS) invariants} introduced in~\cite{Vochten2015} consist of the definition of two FS frames, one for the rotation and one for the translation of the moving body. 
	%The resulting eFS invariants describe the body's rotation and translation, together with the first-order kinematics of the two FS frames. 
	
	The \textit{Denavit Hartenberg inspired Bidirectional} (DHB) invariants \cite{Lee2018} bear a strong relation with the eFS invariants. The DHB invariants are also based on the two FS frames for orientation and translation. But, instead of describing the FS frames' instantaneous first-order kinematics, the DHB invariants describe the relative orientation change between successive FS frames using Denavit-Hartenberg angles.
	
	The eFS and DHB descriptors are \textit{left-invariant}~\cite{Park1995}. They are invariant to changes in the world frame. However, they are not \textit{right-invariant}~\cite{Park1995}, since they depend on a user-selected reference point to define the body's translational velocity. Choosing such a reference point requires a priori knowledge from the user about the points of interest on the moving body. 
	
	Other drawbacks of the eFS and DHB invariants are the following. First, due to the introduction of two decoupled FS frames, the internal kinematics between the rotation and translation of the moving body is lost. For example, the eFS and DHB invariants cannot be used to distinguish screw motions with positive pitch (rotation and translation in the same direction) from ones with negative pitch (rotation and translation in opposite directions).  % Second, the eFS and DHB invariants are not bi-invariant. They require the careful selection of a body reference point to describe the body's translation. Hence, changing the reference point on the moving body will result in different values for the eFS and DHB invariants, although the rigid-body motion itself remains exactly the same. 
	%This is not the case for the bi-invariant ISA invariants and the bi-invariant descriptor ${_f}\boldsymbol{\hat{Y}}(s,\delta s)$. Third, similarly to the ISA invariants, the eFS and DHB invariants provide only a partial description of the body's third-order trajectory shape.
	%Derivatives of the invariants are additionally required to obtain a complete third-order trajectory-shape description.
	Furthermore, the eFS invariants may yield unbounded values~\cite{Vochten2015}. This is not the case for the DHB invariants, since Denavit-hartenberg angles between successive FS frames are always bounded. 
	%Similarity measures derived from the eFS or DHB invariants would inherit above drawbacks. 
	
	%A measure based on the DHB invariants would not be bi-invariant and would not satisfy third-order shape identity. A measure based on the eFS invariants can additionally be unbounded.
	
	The above explained dependency on the reference point can be removed by making use of \textit{screw theory}~\cite{davidson2004robots} such that the descriptor becomes both left- and right-invariant, also referred to as \textit{bi-invariant}~\cite{Park1995}. From the work of Chasles \cite{chasles1830note}, it is known that the kinematics of a moving body can be described in a bi-invariant way by its rotation and translation along the \textit{Instantaneous Screw Axis (ISA)} of the motion. This ISA is a bi-invariant property, since it defines a unique axis of rotation and translation that remains unchanged under reference frame variations.
	The ISA can be represented in multiple ways, including \textit{Pl\"ucker coordinates} and \textit{dual numbers}. 
	
	Based on dual numbers, Stachel~\cite{DualNumberFS2000} derived a bi-invariant moving frame for rigid-body motion based on the ISA and its first-order kinematics. Differentiating this moving frame results in a set of four differential invariants. We refer to these invariants as the \textit{first-order moving frame invariants}. Two of them describe the rotation of the moving frame, while the other two describe its translation.
	
	Veldkamp \cite{veldkamp1967canonical} and Bottema and Roth \cite{bottema1990theoretical} observed that above moving frame can also be used as a local reference frame to characterize the kinematics of any point on the moving body in an invariant way. Modeling the kinematics of these points up to first order requires two invariants, the second order requires four additional invariants, and the third order requires six additional invariants. We refer to these invariants as the \textit{first-order}, \textit{second-order}, and \textit{third-order object invariants}, respectively. However, in \cite{veldkamp1967canonical} and \cite{bottema1990theoretical}, no motion similarity measures were derived from these invariants.  
	
	De Schutter \cite{DeSchutterJoris2010} combined the first-order object invariants and first-order moving frame invariants into a set of six invariants, introducing the \textit{minimal and complete} ISA descriptor. The ISA descriptor is \textit{complete} since it completely describes the evolution of the moving frame and contains just enough information to reproduce the original rigid-body trajectory up to congruence. In \cite{DeSchutter2011} and \cite{Joris2012} a similarity measure for motion recognition applications was derived from the ISA descriptor. To our knowledge, the resulting measure is the only existing local bi-invariant similarity measure that has been applied for motion recognition. However, its recognition accuracy was limited due to the ISA descriptor's high noise sensitivity near singularities in the trajectory. 
	%These singularities include pure translations, pure screw motions, and inflection points. 
	
	In general, all local invariant descriptors have shown a high sensitivity to measurement noise near singularities in the trajectories. Hence, recent works~\cite{WU2008,WU2009,Vochten2015,DeSchutter2011,Joris2012,Lee2018,IntegralInvariants2006,IntegralInvariants2015,vochten2018,vochten2023invariant,boutin2000numerically} aimed to attain higher robustness when calculating invariant descriptors, using techniques such as finite differences~\cite{WU2008,WU2009,boutin2000numerically}, Kalman smoothers~\cite{Joris2012,Vochten2015,DeSchutter2011}, discretized invariants~\cite{Lee2018}, integral invariants~\cite{IntegralInvariants2006, IntegralInvariants2015}, and optimization approaches~\cite{vochten2018, vochten2023invariant}.
	
	Other approaches, such as the Rotation and Relative Velocity (RRV) descriptor approach~\cite{RRV2018} and the Dual Square-Root Function (DSRF) descriptor approach~\cite{DSRF2018} aim to obtain a higher robustness by sacrificing true locality. We refer to this type of descriptors as \textit{global} invariant descriptors.	
	The global RRV and DSRF descriptors obtain left-invariance as follows. The RRV descriptor is based on the definition of one average invariant frame for the entire trajectory. This invariant frame is determined by the left singular vectors of a \mbox{$3\times N$} matrix consisting of a sequence of $N$ rotation axis vectors corresponding to the body's orientation trajectory. This frame is proven to be left-invariant~\cite{RRV2018}. The body's kinematics are then expressed in this frame to obtain a left-invariant trajectory descriptor. The DRSF approach obtains left-invariance by using a global world frame orientation alignment action before comparing two trajectory descriptors.
	However, similarly to the eFS and DHB descriptor approaches, both the RRV and DSRF approaches lack right-invariance, since they also depend on a user-selected reference point to define the body's translational velocity.
	
	Despite their robustness, global descriptors exhibit an important drawback: the loss of true locality. This limitation reduces their effectiveness when applied to poorly segmented data. It also makes them less suited for trajectory segmentation based on local features. Additionally, using global descriptors can result in higher computational overhead when combined with alignment algorithms like DTW over a sliding window for real-time applications. 
	%Specifically, consider two trajectory segments. One represents a static reference, while the other represents a part (determined by a sliding window) of a continuous real-time trajectory. When this window shifts, updating the DTW cost between these segments is relativey inefficient in case of global descriptors compared to local descriptors. 
	Specifically, for each window shift, the entire global descriptor for the current window has to be recalculated, along with all corresponding local pairwise DTW costs. In contrast, local descriptors enable more efficient DTW cost updates since they allow to reuse most of the local pairwise DTW costs from the previous window. This reduces the computational overhead, making local descriptors more efficient for real-time applications.

	\subsubsection*{Comparison of existing approaches} 
	The key properties of similarity measurement approaches based on existing invariant descriptors are summarized in Table~\ref{tab:overview_literature}. For comparison, the properties of the proposed BILTS$^+$ measure are also included.
	
	\begin{table*}[t]
		\centering
		\caption{Properties of Invariant Object Motion Similarity Measurement Approaches}
		\resizebox{0.75\textwidth}{!}{
			\begin{tabular}{lcccccccc}
				\toprule
				\textbf{Approach} &
				\multicolumn{2}{c}{\textbf{The measure is:}}
				& \multicolumn{2}{c}{\textbf{The measure is robust to:}} & \multicolumn{4}{c}{\textbf{The descriptor is robust to noise for the special motions:}} \\
				\cmidrule(lr){2-3} \cmidrule(lr){4-5} \cmidrule(lr){6-9}
				& \multirow{2}{*}{\textbf{local}} & \multirow{2}{*}{\textbf{bounded}} & \textbf{world frame} & \textbf{body frame}  & \textbf{pure rotation} & \textbf{screw motion} & \textbf{pure rotation} & \textbf{pure} \\
				&  &  & \textbf{variations} & \textbf{variations} & \textbf{(fixed axis)} & \textbf{(fixed axis)} & \textbf{(pivoting axis)} &  \textbf{translation}\\
				\midrule
				RRV~\cite{RRV2018}       & \xmark & \cmark & \cmark & \xmark & \xmark & \xmark & \xmark & \xmark \\
				DSRF~\cite{DSRF2018}      & \xmark & \cmark & \cmark & \xmark & \xmark & \cmark & \xmark & \cmark \\
				DHB~\cite{Lee2018}       & \cmark & \cmark & \cmark & \xmark & \xmark & \xmark & \xmark & \xmark \\
				eFS~\cite{Vochten2015}       & \cmark & \xmark & \cmark & \xmark & \xmark & \xmark & \xmark & \xmark \\
				ISA~\cite{DeSchutterJoris2010}       & \cmark & \xmark & \cmark & \cmark & \xmark & \xmark & \cmark & \xmark \\ \midrule
				\textbf{BILTS$^+$} & \cmark & \cmark & \cmark & \cmark & \cmark & \cmark & \cmark & \cmark \\
				\bottomrule
		\end{tabular}}
		\vspace{-7pt} 
		\label{tab:overview_literature} \\
	\end{table*}
	
	A shared challenge among these approaches is their high noise sensitivity near singularities in the trajectory. These singularities can be understood as trajectory segments characterized by special motion types that do not excite sufficient degrees of freedom to completely and uniquely determine the approach's invariant descriptor. A few special motion types are listed in Table~\ref{tab:overview_literature}. Columns six through nine indicate each approach's robustness in handling these special motion types.
	
	Most of the existing approaches cannot robustly deal with pure rotations along a fixed rotation axis or pivoting axis, where the original body reference point lies on this axis. The RRV and DSRF approaches normalize the reference point trajectory (which is a stationary point trajectory for these special cases) to unit length, which introduces ambiguity and severe noise sensitivity in the rescaled position trajectory. Also, due to this stationary point trajectory, the FS frame for translation (used within the eFS and DHB approaches) is ill-defined in these cases. Only the ISA approach can robustly deal with pure rotations about a pivoting rotation axis.
	
	Most approaches cannot robustly deal with pure screw motions and pure translations. Specifically, the functional frame of the RRV descriptor, the FS frame for rotation, and the moving frame of the ISA descriptor are completely ill-defined for pure translations. These frames are also partially ill-defined for rotations about a fixed axis, as in this case, only the first axis of the frame remains well defined. 
	% Only the DSRF approach can robustly deal with these special motion types, as the rescaling of the position trajectory remains well defined in these cases.
	
	To conclude, Table~\ref{tab:overview_literature} shows that the ISA measure is the only existing similarity measure that obtains full left- and right-invariance (bi-invariance), resulting in a high robustness to variations in both the world and body reference frames. However, it suffers from noise sensitivity and unbounded values when dealing with singularities. 
	To adress this lack of robustness, we introduce a novel bi-invariant similarity measure (BILTS) and propose regularization actions (BILTS$^+$) to robustly deal with singularities. The properties of the proposed regularized measure are listed in Table~\ref{tab:overview_literature}.
	
	\subsection{Paper contributions}

	In this paper, we introduce the novel \textit{Bi-Invariant Local Trajectory-Shape Similarity} (BILTS) measure for rigid-body motion. This measure quantifies the similarity between two trajectory shapes as a single scalar value in a bi-invariant manner. To introduce the BILTS measure, we first devise a bi-invariant trajectory-shape descriptor by expressing second-order Taylor series approximation of the body's kinematics in the bi-invariant frame of~\cite{veldkamp1967canonical,bottema1990theoretical,DeSchutterJoris2010}. We then define the dissimilarity between two trajectory shapes as the square root of the weighted sum of the squared elementwise differences between their descriptor elements. For this weighting, we identify two minimally necessary weighting factors (a geometric scale and a progress scale).
	
	Furthermore, we derive analytical relations between the BILTS measure and other measures based on existing invariant descriptors. Compared to these measures, only the BILTS measure possesses both bi-invariance and boundedness. It is also the only one that captures third-order shape identity.  
	
	To facilitate practical implementation, we also devise a discretized version of the BILTS measure which can be computed from discrete-time rigid-body trajectories without requiring explicit estimation of higher-order trajectory derivatives. Lastly, we introduce regularization (BILTS$^+$) to increase the BILTS measure's robustness to noise near singularities.
	
	We validate the BILTS and BILTS$^+$ measures through object motion recognition experiments. Results confirm that the BILTS measure, guaranteed to be bounded, outperforms the ISA measure in recognizing synthetic rigid-body motions that are rich in singularities. Across multiple datasets, the regularized BILTS$^+$ further demonstrated exceptional robustness to singularities, achieved the highest average recognition rate, and demonstrated the lowest sensitivity to contextual variations. These findings establish the superiority of BILTS$^+$ over other object motion similarity measures~\cite{DeSchutter2011,Joris2012,Vochten2015,Lee2018,DSRF2018,RRV2018}.
	
	Lastly, we discuss other uses for the BILTS$^+$ measure, supported by experimental proofs of concept. These include the analysis of a tool's motion along a 3D contour for trajectory segmentation purposes, and a real-time demonstration of gesture recognition for human-robot interaction purposes.
	
	%outline
	\begin{comment}
	\subsection{Paper outline}
	This paper is organized as follows. Section \ref{sec:prelim} reviews mathematical preliminaries. Section \ref{sec:descriptor} introduces the novel continuous-time BILTS measure and derives properties and relations with other approaches.  Section~\ref{sec:practical} introduces a numerical \textit{extended QR (eQR) decomposition} algorithm to compute the bi-invariant descriptor of the BILTS measure. It then introduces the discretized BILTS measure and regularization actions to improve its robustness to singularities. The proposed BILTS measure is validated for the recognition of object-manipulation tasks performed in diverse contexts in Section~\ref{sec:recognition}. Section \ref{sec:discussion_conclusion} provides a discussion supported by additional experimental proofs of concept, and a conclusion.
	\end{comment}

	\section{Mathematical Preliminaries and Notation}
	\label{sec:prelim}
	
	%This section introduces the necessary background and notation on rigid-body kinematics.
	
	\subsection{Fundamentals of Rigid-Body Trajectories}
	A \textit{rigid body} is an object that cannot be deformed. The body's location in space is commonly described by the relative position and orientation (together: \textit{pose}) between two Cartesian reference frames. One frame $\{b\}$ is rigidly attached to the body (\textit{body frame}) while the other frame $\{w\}$ is fixed with respect to another body, e.g. the world (\textit{spatial frame}).
	
	From Lie group theory~\cite{murray1994mathematical}, it is known that the relative pose in 3D of frame $\{b\}$ with respect to frame $\{w\}$ can be written using a $4\times4$ homogeneous transformation matrix $\tf{w}{b}$, consisting of the \textit{rotation matrix} $\boldsymbol{Q}$ and \textit{position vector} $\boldsymbol{p}$:
	\begin{equation}
		\tf{w}{b} =  %\left.  
		\begin{bmatrix}
			\boldsymbol{Q} & \boldsymbol{p} \\
			0        & 1
		\end{bmatrix},~\text{with}~ %\right|
		\begin{matrix}
			\boldsymbol{Q} \in \mathbb{R}^{3\times3},
			\boldsymbol{p} \in \mathbb{R}^3, \\
			~~~\boldsymbol{Q}^T \boldsymbol{Q} = \boldsymbol{I},~
			\det(\boldsymbol{Q})=1.
		\end{matrix}
	\end{equation}
	The set of all such homogeneous transformation matrices $\tf{w}{b}$ is referred to as the \textit{special Euclidean group}~$SE(3)$~\cite{murray1994mathematical}. 
	
	A rigid-body trajectory $\tf{w}{b}(s)$ describes the evolution of $\tf{w}{b}$ as a function of a \textit{progress variable} $s$. Choosing $s$ to correspond to the elapsed time yields a \textit{temporal rigid-body trajectory}. Choosing $s$ to correspond to a \textit{geometric progress}, such as the traversed angle of the moving body~\cite{Roth2005}, or the traversed arclength along the path traced by the reference point on the moving body \cite{RRV2018,DSRF2018} yields a \textit{geometric trajectory}. The advantage of geometric trajectories is that their higher-order kinematics (velocity, acceleration, etc.) are fully determined by the trajectory's geometry, independent of how fast or slow the trajectory is traversed over time. This allows motion to be studied purely based on spatial geometry.
	
	In the remainder of this paper, the argument $(s)$ referencing the progress instance will be omitted from variables such as $\tf{w}{b}(s)$ when clear from the context.
	
	\subsection{Screw twists as a basis for trajectory-shape descriptions}
	
	The shape of a trajectory $\tf{w}{b}(s)$ at a certain progress value $s$ is defined by the trajectory's higher-order derivatives $\tf{w}{b}'(s)$, $\tf{w}{b}''(s)$, $\tf{w}{b}'''(s)$, etc. The derivative $\tf{w}{b}'(s)$ is considered an element of the tangent space to $SE(3)$. The tangent space at group identity is formally known as the \textit{Lie algebra} of the Lie group. The Lie algebra of $SE(3)$, denoted as $se(3)$, is characterized by $4\times4$ matrices with the following structure:
	\begin{equation}
		\skewsym{\tw}=
		\begin{bmatrix}
			\skewsym{\boldsymbol{\omega}} & \boldsymbol{v} \\
			0                & 0
		\end{bmatrix},\text{~with~} %\left|
		\boldsymbol{\omega}  \in \mathbb{R}^{3},
		\boldsymbol{v} \in \mathbb{R}^3,
		%\right.
		%		\skewsym{\omega}^T = - \skewsym{\omega}
	\end{equation}
	having the matrix commutator as the Lie bracket.
	The skew-symmetric matrix $\skewsym{\boldsymbol{\omega}}$ is associated with the rotational vector $\boldsymbol{\omega}$ as follows:
	\begin{equation}
		\skewsym{\boldsymbol{\omega}} = \small
		\begin{bmatrix}
			0         & -\omega_z & \omega_y  \\
			\omega_z  & 0         & -\omega_x \\
			-\omega_y & \omega_x  & 0
		\end{bmatrix}.
	\end{equation}
	
\begin{figure}[t]
		\centering
		\includegraphics[width=0.7\linewidth]{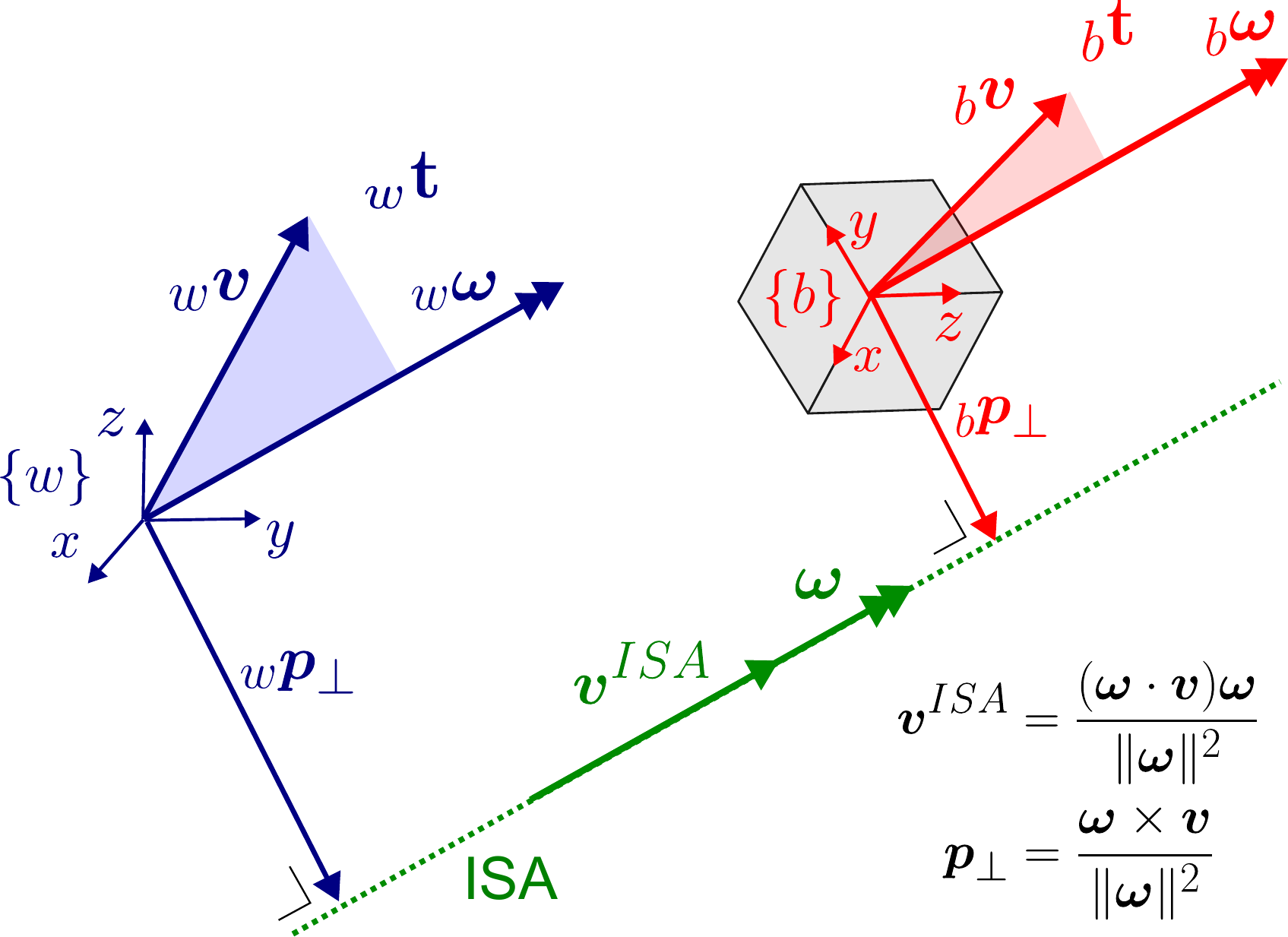}
		\caption{Visualization of the spatial screw twist $_w\tw{}$, the body-fixed screw twist $_b\tw{}$, and the Instantaneous Screw Axis (ISA) of the rigid-body motion. Rotational velocity vectors are indicated by double-headed arrows, while translational velocity vectors are represented by single-headed arrows. The translational velocity parallel to the ISA, $\boldsymbol{v}^{ISA}$, and the location of the point on the ISA closest to the origin of the reference frame, ${_w}\boldsymbol{p}_\perp$ or ${_b}\boldsymbol{p}_\perp$, can be calculated from the screw twists $_w\tw{}$ or $_b\tw{}$, respectively. Corresponding calculation formulas are provided at the lower right of the figure.}	
		\label{fig:twist_and_ISA}
	\end{figure}

	The matrix $\skewsym{\tw}$ is characterized by six independent coordinates in the rotation vector $\boldsymbol{\omega}$ and translation vector $\boldsymbol{v}$. We introduce the notation $\tw= \small \Big(\begin{array}{c} \boldsymbol{\omega} \\ \boldsymbol{v} \end{array}\Big)$ and refer to this $6\times1$ vector as the \textit{screw twist}.
	The notation~$_b\tw{}$ represents the \emph{body-fixed} screw twist. 
	It consists of the rotational velocity vector ${_b}\boldsymbol{\omega}$ of $\{b\}$ and the translational velocity vector ${_b}\boldsymbol{v}$ of the origin of $\{b\}$, both expressed in $\{b\}$.
	The notation~$_w\tw{}$ represents the \emph{spatial} screw twist. It consists of the rotational velocity vector ${_w}\boldsymbol{\omega}$ of $\{b\}$ and the translational velocity vector ${_w}\boldsymbol{v}$ of a reference point rigidly attached to $\{b\}$ which instantaneously coincides with the origin of $\{w\}$, both expressed in $\{w\}$. 
	The twists, $_w\tw{}$ and $_b\tw{}$, are visualized in Figure~\ref{fig:twist_and_ISA}. 
	% The spatial screw twist $_w\tw{}$ and body-fixed screw twist $_b\tw{}$ are visualized in Figure~\ref{fig:twist_and_ISA}. 
	
	Any rigid-body motion can be represented by a rotation about and a translation parallel to the Instantaneous Screw Axis (ISA) of the motion~\cite{chasles1830note}. The twists $_b\tw{}$ and $_w\tw{}$ are referred to as `screw twists', since they contain the necessary parameters for identifying this ISA of the motion (see Fig.~\ref{fig:twist_and_ISA}).
	
	The body-fixed screw twist~$_b\tw$ is related to the derivative $\tf{w}{b}' = \frac{d}{ds}(  \tf{w}{b})$ as follows:
	\begin{equation}
		\skewsym{_b\tw} = (\tf{w}{b}^{-1}) \ \tf{w}{b}'.
		\label{eq:movingframe_body}
	\end{equation}
	The body-fixed screw twist $_b\tw$ is not influenced by the choice for the spatial frame $\{w\}$ attached to the world. Consider choosing another frame $\{a\}$ rigidly attached to $\{w\}$ as the spatial frame, with the relation between $\{w\}$ and $\{a\}$ described by the constant homogeneous matrix~$\tf{a}{w}$, such that $\tf{a}{b} = \tf{a}{w}\tf{w}{b}$. Calculating the body-fixed screw twist $_b\tw$ from $\tf{a}{b}$ and $\tf{a}{b}'$ then results in:
	\begin{align*}
		\skewsym{_b\tw} &= (\tf{a}{b})^{-1} \ \tf{a}{b}' ,\\
		&= (\tf{a}{w}\tf{w}{b})^{-1} \ (\tf{a}{w}\tf{w}{b})' ,\\
		&= (\tf{w}{b}^{-1}\tf{a}{w}^{-1}) \ (\tf{a}{w}\tf{w}{b}'), \\
		&= (\tf{w}{b})^{-1}\tf{w}{b}'.
	\end{align*}
	Hence, changing the spatial frame does not alter the values in ${_b}\tw$. The body-fixed screw twist $_b\tw$ is said to be a \emph{left-invariant} representation of the motion \cite{murray1994mathematical}.
	%\cite{belta2001motion}
	
	The spatial screw twist $_w\tw$ is related to the derivative $\tf{w}{b}'$:
	\begin{equation}
		\skewsym{_w\tw} = \tf{w}{b}' (\tf{w}{b})^{-1}.
		\label{eq:fixedframe}
	\end{equation}
	Changing the body frame attached to the moving body does not alter the values in ${_w}\tw$. This can be proven in a manner analogous to the proof for $_b\tw$. The spatial screw twist $_w\tw$ is said to be a \emph{right-invariant} representation of the motion \cite{murray1994mathematical}.
	
	As previously mentioned, the shape of the trajectory $\tf{w}{b}$ at a certain progress value $s$ is defined by the trajectory's higher-order derivatives $\tf{w}{b}'(s)$, $\tf{w}{b}''(s)$, $\tf{w}{b}'''(s)$, etc.
	As shown in~\eqref{eq:movingframe_body} and~\eqref{eq:fixedframe}, the trajectory derivative $\tf{w}{b}'$ is described in a left- or right-invariant manner by $_b\tw$ or $_w\tw$, respectively. Hence, we can describe the trajectory shape of $\tf{w}{b}$ in a left- or right-invariant manner through the respective screw twist, $_b\tw$ or $_w\tw$, and their higher-order derivatives. 
	
	\subsection{Screw twist transformation matrix}
	Using \eqref{eq:movingframe_body} and \eqref{eq:fixedframe}, transforming the twist $_b\tw$ to $_w\tw$ (i.e. transforming the reference frame from $\{b\}$ to $\{w\}$) is written as the following \textit{adjoint action}:
	\begin{equation}
		\skewsym{_w\tw}  =  \tf{w}{b} \, \skewsym{_b\tw} \, (\tf{w}{b})^{-1},
		\label{eq:skew_similaritytf}
	\end{equation}
	which can directly be written in function of the screw twists:
	\begin{equation}
		_w\tw = \boldsymbol{S}(\tf{w}{b}) ~_b\tw,
		\label{eq:screwtf_twist}
	\end{equation}
	using the $6\times6$ \textit{screw transformation matrix} $S(T)$:
	\begin{equation}
		\boldsymbol{S}(\tf{w}{b}) = \begin{bmatrix}
			\boldsymbol{Q}                    & \boldsymbol{0} \\
			\skewsym{\boldsymbol{p}} \boldsymbol{Q} & \boldsymbol{Q}
		\end{bmatrix}.
		\label{eq:screwtf}
	\end{equation}
	The matrix $\boldsymbol{S}(\tf{w}{b})$ is also referred to as the \textit{adjoint representation} of $\tf{w}{b}$.
	By taking the derivative of the above and further algebraic manipulation, the derivative $\boldsymbol{S}'(\tf{w}{b})$ of a screw twist transformation $\boldsymbol{S}(\tf{w}{b})$ can be written as:
	\begin{equation}
		\boldsymbol{S}'(\tf{w}{b}) = \boldsymbol{S}(\tf{w}{b}) \skewskewsym{{_b\tw}},
		\label{eq:deriv_screw_transform}
	\end{equation}
	with $\skewskewsym{_b\tw}$ defined as:
	\begin{equation}
		\skewskewsym{_b\tw} = \begin{bmatrix}
			\skewsym{\boldsymbol{\omega}}   & \boldsymbol{0}         \\
			\skewsym{\boldsymbol{v}} & \skewsym{\boldsymbol{\omega}}
		\end{bmatrix}.
		\label{eq:spatialcrossprod}
	\end{equation}

	\section{Continuous-time BILTS measure}
	\label{sec:descriptor}
	
	This section introduces the continuous-time formulation of the novel Bi-Invariant Local Trajectory-shape Similarity (BILTS) measure, which quantifies the difference in trajectory shape as a single scalar value in a bi-invariant manner. 
	
	This section is built up as follows. In Subsection~\ref{sec:theo_BILTS}, a right-invariant descriptor of the trajectory shape at progress instance $s$ is first derived from second-order Taylor series expansions of the screw twist $_w\tw(s)$. Afterwards, a bi-invariant descriptor is derived by introducing a functional bi-invariant frame $\{f\}$. The BILTS measure between two trajectory-shape descriptors is then introduced as the square root of the weighted sum of the squared differences between corresponding descriptor elements. Properties of the BILTS measure are listed in Subsection~\ref{sec:properties}. Relations between the BILTS measure and other invariant similarity measures based on existing trajectory-shape descriptors are derived in Subsection~\ref{sec:relation}. 
	
	\subsection{Derivation of the Continuous-time BILTS measure}
	\label{sec:theo_BILTS}
	
\begin{figure}[t]
		\centering
		\includegraphics[width=0.6\columnwidth]{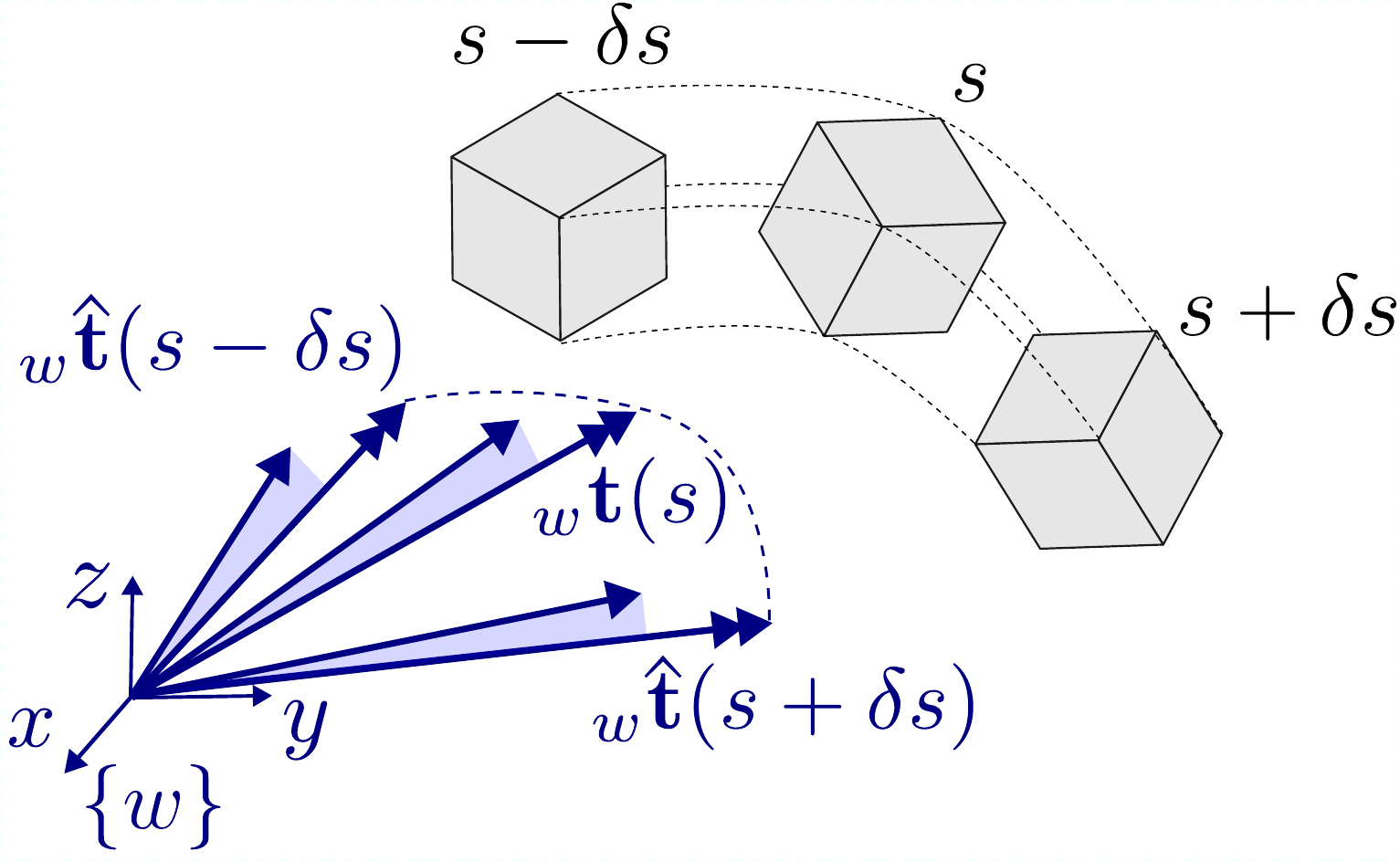}
		\caption{Visualization of the right-invariant shape descriptor consisting of three spatial screw twists expressed in the spatial frame $\{w\}$. These twists are right-invariant, i.e. they are independent on the location and orientation of the body frame $\{b\}$.}	
		\vspace{-10pt}
		\label{fig:taylor1}
	\end{figure}

	In this section, we assume continuous-time trajectories. Furthermore, we assume that the twist $_w\tw(s)$ and its higher-order derivatives along the trajectory are known. Using these derivatives, we can perform second-order Taylor series approximations of the twists $_w\tw(s-\delta s)$ and $_w\tw(s+\delta s)$ with $\delta s$ a chosen progress increment:
	\begin{align}
		{_w\hat{\tw}}(s-\delta s) &= {_w\tw}(s) -  {_w\tw}'(s) \delta s +  {_w\tw}''(s) \frac{\delta s^2}{2}, \label{eq:TS1} \\
		{_w\hat{\tw}}(s+\delta s) &= {_w\tw}(s) +  {_w\tw}'(s) \delta s +  {_w\tw}''(s) \frac{\delta s^2}{2}.
		\label{eq:TS2}
	\end{align}
	In \eqref{eq:TS1} and \eqref{eq:TS2}, we introduced the hat notation $\hat~$ to explicitly denote that ${_w\hat{\tw}}(s-\delta s)$ and ${_w\hat{\tw}}(s+\delta s)$ are second-order approximations of ${_w\tw}(s-\delta s)$ and ${_w\tw}(s+\delta s)$, respectively. The three twists ${_w\hat{\tw}}(s-\delta s)$, ${_w\tw}(s)$, and ${_w\hat{\tw}}(s+\delta s)$ can be written more compactly by stacking them into the $6\times3$ matrix $_w\boldsymbol{\hat{Y}}(s,\delta s)= \begin{bmatrix}{_w\hat{\tw}}(s-\delta s) & {_w\tw}(s) & {_w\hat{\tw}}(s+\delta s) \end{bmatrix}$. Similarly, the spatial screw twist and its derivatives can be stacked into the $6\times3$ matrix $_w\boldsymbol{X}(s)= \begin{bmatrix} {_w\tw}(s) & {_w\tw}'(s) & {_w\tw}''(s) \end{bmatrix}$.  
	Equations \eqref{eq:TS1} and \eqref{eq:TS2} can then be written as:
	\begin{equation}
		_w\boldsymbol{\hat{Y}}(s,\delta s) = ~_w\boldsymbol{X}(s) \ \boldsymbol{C}(\delta s)
		\label{eq:Ad_def},
	\end{equation}
	where $\boldsymbol{C}(\delta s)$ contains the Taylor series coefficients related to the progress increment~$\delta s$:
	\begin{equation}
		\label{eq:taylor_matrix}
		\boldsymbol{C}(\delta s) = \begin{bmatrix} 1 & 1 & 1 \\
			-\delta s & 0 & \delta s \\
			\frac{\delta s^2}{2} & 0 & \frac{\delta s^2}{2} \end{bmatrix}.
	\end{equation}
	The matrix $\boldsymbol{C}(\delta s)$ is an invertible matrix because we used second-order Taylor series approximations in \eqref{eq:TS1} and \eqref{eq:TS2}. Since $\boldsymbol{C}(\delta s)$ is invertible for all $\delta s \neq 0$, we can retrieve the twist derivatives in $_w\boldsymbol{X}(s)$ from $_w\boldsymbol{\hat{Y}}(s,\delta s)$ and $\boldsymbol{C}(\delta s)$. Hence, the matrix $_w\boldsymbol{\hat{Y}}(s,\delta s)$ completely and uniquely describes the twist $_w\tw(s)$ and its first- and second-order derivatives for a given progress increment $\delta s$. 
	%Therefore, the matrix $_w\hat{Y}(s,\delta s)$ describes first-, second-, and third-order derivatives of the trajectory $\tf{w}{b}(s)$. 
	We refer to $_w\boldsymbol{\hat{Y}}(s,\delta s)$ as a \textit{third-order trajectory-shape descriptor} (see Fig. \ref{fig:taylor1}).
	
	The progress increment $\delta s$ in $_w\boldsymbol{\hat{Y}}(s,\delta s)$ can be intepreted as a physically-inspired parameter, controlling the sensitivity of the numbers in $_w\boldsymbol{\hat{Y}}(s,\delta s)$ to variations in the first- and second-order twist derivatives. 
	
	The descriptor $_w\boldsymbol{\hat{Y}}(s,\delta s)$ describes the spatial screw twist and its derivatives, which are \textit{right-invariant}. Consequently, $_w\boldsymbol{\hat{Y}}(s,\delta s)$ inherits this right-invariance.
	% and is therefore independent of the orientation and position of the body frame $\refframe{b}$.
	However, it is not \textit{left-invariant}. Consequently, it remains dependent on the orientation and location of the spatial frame $\{w\}$. 
	
\begin{figure}[t]
		\centering
		\subfloat[]{\includegraphics[width=0.33\linewidth]{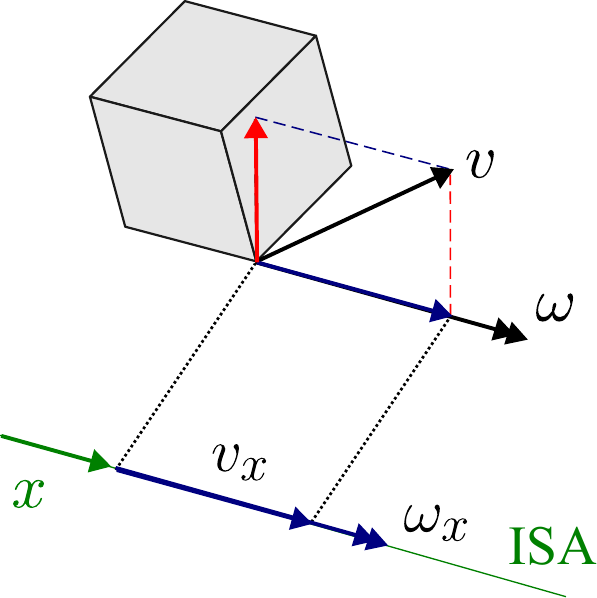} 
		\label{fig:movingframe1}} \subfloat[]{\includegraphics[width=0.33\linewidth]{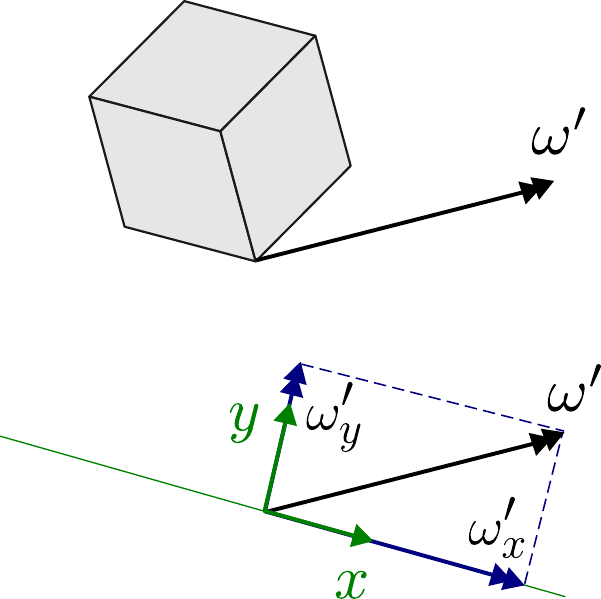}
			\label{fig:movingframe2}}
		\subfloat[]{\includegraphics[width=0.31\linewidth]{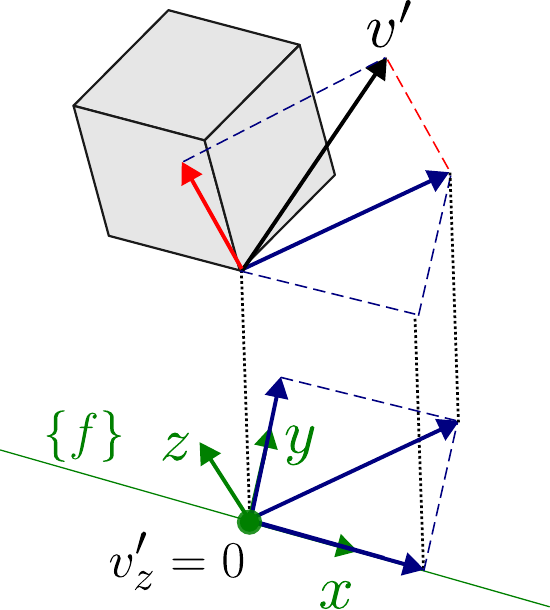}
			\label{fig:movingframe3}}
		\caption{Construction of the functional bi-invariant frame $\{f\}$.}
		\label{fig:movingframe}
	\end{figure}

	To additionally obtain left-invariance, we need to remove the dependency on the spatial frame $\{w\}$. We can do this by changing the spatial frame $\{w\}$ to a functional and bi-invariant frame $\{f\}$. To this end, we utilize the functional and bi-invariant frame for rigid-body motion introduced in \cite{veldkamp1967canonical,bottema1990theoretical,DeSchutterJoris2010,vochten2023invariant}. This frame is functional, since it is completely determined by the motion itself, specifically through the motion's Instantaneous Screw Axis (ISA) and its first-order kinematics. 
	%This ISA is a bi-invariant property of the motion, since it defines a unique axis of rotation and translation that remains unchanged under reference frame transformations~\cite{chasles1830note}. Hence, the functional frame, derived from the ISA and its first-order kinematics, inherits this bi-invariance. 
	While the works \cite{veldkamp1967canonical,bottema1990theoretical,DeSchutterJoris2010,vochten2023invariant} describe the same functional bi-invariant frame based on the ISA, they use different conventions for the sign and order of the frame's axes. Following the convention of \cite{vochten2023invariant}, this frame is built in the following three steps:
	
	(1) Align the $x$-axis precisely with the ISA of the motion. Due to this alignment, the $y$- and $z$-components of both the rotational and translational velocity vectors become zero (see Figure~\ref{fig:movingframe1}). The sign of the $x$-axis is chosen such that $\omega_x>0$.
	
	(2) Orient the $y$-axis normal to the $x$-axis such that the $z$-component of the rotational acceleration vector $\boldsymbol{\omega}'$ becomes zero. Hence, the $x$- and $y$-axis span the osculating plane of the rotational velocity vector (see Figure~\ref{fig:movingframe2}). The sign of the $y$-axis is chosen such that $\omega'_y>0$.
	
	(3) Position the frame's origin precisely at the \textit{striction point}~\cite{bottema1990theoretical,veldkamp1967canonical} on the ISA.  Due to this positioning, the $z$-component of the translational acceleration vector $\boldsymbol{v}'$ becomes zero (see Fig.~\ref{fig:movingframe3}). 
	
	This procedure completely determines the pose of the frame $\{f\}$, since the $z$-axis follows from the $x$- and $y$-axis through right-handed orthogonality. The relative pose of $\{f\}$ with respect to the original spatial frame $\{w\}$ is denoted by $\tf{w}{f}$.
	
\begin{figure}[t]
		\centering
		\includegraphics[width=0.5\columnwidth]{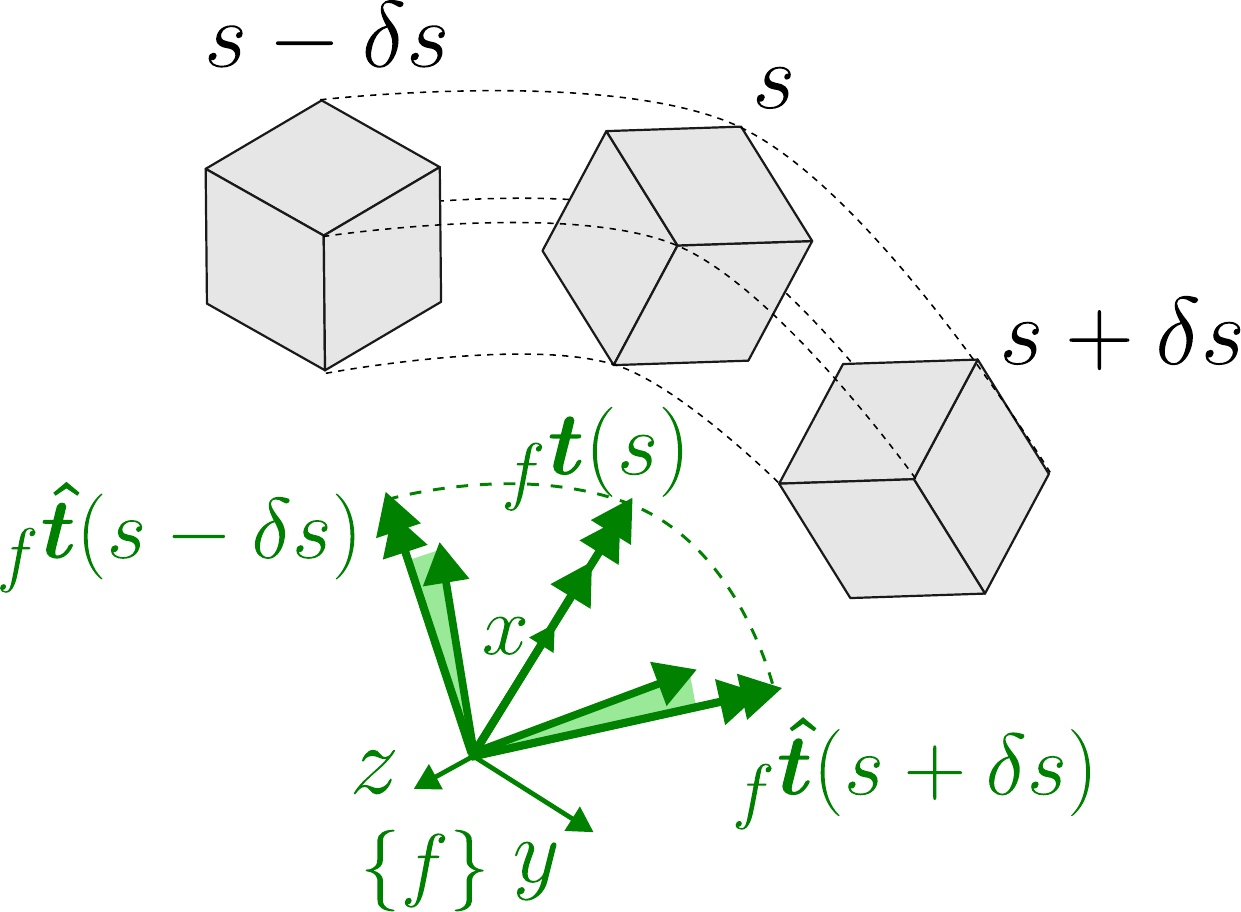}
		\caption{Illustration of the bi-invariant shape descriptor consisting of the same three twists in Figure~\ref{fig:taylor1}, but now expressed in the functional bi-invariant frame~$\{f\}$. These twists are additionally left-invariant, hence they are also independent on the location and orientation of the spatial frame $\{w\}$.}	
		\vspace{-5pt}
		\label{fig:taylor2}
	\end{figure}

	Once the pose of the frame $\{f\}$ is determined, we can change the spatial frame $\{w\}$ of $_w\boldsymbol{X}(s)$ to the frame $\refframe{f}$ using the screw transformation matrix $\boldsymbol{S}$ in \eqref{eq:screwtf_twist}:
	\begin{equation}
		_f\boldsymbol{X}(s)= \boldsymbol{S}^{-1}(\tf{w}{f}) ~_w\boldsymbol{X}(s).
		\label{eq:matrix_decomp0}
	\end{equation} 
	The numbers in $_f\boldsymbol{X}(s) = \left[ {_f\tw} ~~ {_f\tw}' ~~ {_f\tw}'' \right] $ are bi-invariant, since $\{f\}$ serves as a bi-invariant reference frame. Due to the previously described convention for the axes of $\{f\}$, $_f\boldsymbol{X}(s)$ obtains a twice upper-triangular form. Equation~\eqref{eq:matrix_decomp0} can hence be interpreted as a \textit{decomposition} or \textit{factorization} of $_w\boldsymbol{X}(s)$ into the product of a screw transformation matrix $\boldsymbol{S}$ and a twice upper-triangular matrix $\boldsymbol{R}(s)$:
	\begin{equation}
		_w\boldsymbol{X}(s) = \boldsymbol{S}(\tf{w}{f}) \ \boldsymbol{R}(s),
		\label{eq:matrix_decomp}
	\end{equation} 
	with
	\begin{equation}
		\label{eq:matrix_decomp2}
		\small 
		\boldsymbol{R}(s) = \begin{bmatrix} r_{11} & r_{12} & r_{13} \\	0 & r_{22} & r_{23} \\
			0 & 0 & r_{33} \\
			r_{41} & r_{42} & r_{43} \\	
			0 & r_{52} & r_{53} \\
			0 & 0 & r_{63} \end{bmatrix}, ~~~ \text{and} ~~~ \Bigg\{ 
		\begin{array}{c} 
			r_{11} > 0 \\ r_{22} > 0 
		\end{array} .
	\end{equation} 
	The bi-invariant numbers $r_{ij}$ in \eqref{eq:matrix_decomp2} were already discovered in the works of \cite{bottema1990theoretical,veldkamp1967canonical}. Nevertheless, we have now shown that these invariants naturally align with a twice upper-triangular form for $\boldsymbol{R}(s)$. Applying this matrix decomposition in \eqref{eq:Ad_def} results in:
	\begin{equation}
		_w\boldsymbol{\hat{Y}}(s,\delta s) =  \boldsymbol{S}(\tf{w}{f}) \ \boldsymbol{R}(s) \ \boldsymbol{C}(\delta s).
		\label{eq:matrix_decomp_discrete}
	\end{equation}
	Pre-multiplying both sides of \eqref{eq:matrix_decomp_discrete} by $\boldsymbol{S}^{-1}(\tf{w}{f})$ results in:
	\begin{equation}
		\boldsymbol{S}^{-1}(\tf{w}{f}) \ _w\boldsymbol{\hat{Y}}(s,\delta s) = \boldsymbol{R}(s) \ \boldsymbol{C}(\delta s).
	\end{equation}
	Using $\boldsymbol{S}^{-1}(\tf{w}{f}) \ _w\boldsymbol{\hat{Y}}(s,\delta s) = {_f}\boldsymbol{\hat{Y}}(s,\delta s)$ then results in:
	\begin{equation}
		_f\boldsymbol{\hat{Y}}(s,\delta s) = \boldsymbol{R}(s) \ \boldsymbol{C}(\delta s).
		\label{eq:def_Y}
	\end{equation}
	Since $\boldsymbol{R}(s)$ is bi-invariant, the trajectory-shape descriptor \mbox{$_f\boldsymbol{\hat{Y}}(s,\delta s)$} inherits this bi-invariance. Also, since the twelve elements in $\boldsymbol{R}(s)$ can vary freely, there are 12 degrees of freedom (DOFs) on the right-hand side of \eqref{eq:def_Y} for a given $\delta s$. Consequently, the bi-invariant descriptor $_f\boldsymbol{\hat{Y}}(s,\delta s)$ also encodes 12~DOFs. It has the following form:
	\begin{align}
		_f\boldsymbol{\hat{Y}}(s,\delta s) &= \left[ {_f}\hat{\tw}^- \hspace{5pt} {_f}\tw \hspace{8pt} {_f}\hat{\tw}^+ \right]\\
		\label{eq:BILTS_B_matrix}
		&= 
			\begin{bmatrix}
				\omega_x^- & \omega_x & \omega_x^+ \\
				\omega_y^- & 0 & \omega_y^+ \\
				\omega_z^{-} & 0 & \omega_z^+\\
				v_x^- & v_x & v_x^+ \\
				v_y^- & 0 & v_y^+ \\
				v_z^- & 0 & v_z^+
			\end{bmatrix}
			~\text{with} ~ \left\{ \begin{array}{c} \hspace{17pt}\omega_x > 0 \\ 
				\hspace{-10pt}\omega_y^+-\omega_y^- > 0 \\ 
				\hspace{23pt}\omega_z^- = \omega_z^+ \\ 
				\hspace{23pt}v_z^- = v_z^+
			\end{array} \right.,
	\end{align} 
	and where the $-$ and $+$ superscripts indicate the second-order Taylor series approximation at \mbox{$s-\delta s$} and $s+\delta s$, respectively. The descriptor $_f\boldsymbol{\hat{Y}}(s,\delta s)$ describes the third-order trajectory shape of $\tf{w}{b}(s)$ in a bi-invariant way. Due to this bi-invariance, changing the body frame $\{b\}$ or world frame $\{w\}$ does not alter the values in $_f\boldsymbol{\hat{Y}}(s,\delta s)$. The bi-invariant descriptor $_f\boldsymbol{\hat{Y}}(s,\delta s)$ is visualized in Figure~\ref{fig:taylor2}. 
	
	The dissimilarity between two rigid-body trajectory shapes at given progress $s$ can now be defined in a bi-invariant way using the trajectory-shape descriptor $_f\boldsymbol{\hat{Y}}(s,\delta s)$. We define the dissimilarity $d$ between two trajectory shapes, $_f\boldsymbol{\hat{Y}}_1(s,\delta s)$ and $_f\boldsymbol{\hat{Y}}_2(s,\delta s)$, as the square root of the weighted sum of the squared elementwise differences between the shapes: 
	\begin{align}
		d ~\defeq~ & 
		\left\| _f\boldsymbol{\hat{Y}}_1(s,\delta s) - {_f}\boldsymbol{\hat{Y}}_2(s,\delta s) \right\|_{\boldsymbol{W}},
		\label{eq:localmetric0}
	\end{align}
	where $\Vert ... \Vert_W$ represents a weighted Frobenius norm with $\boldsymbol{W} = \text{diag}\left(L^2,L^2,L^2,1,1,1\right)$.
	The parameter $L$ has units [length] and ensures an appropriate weighting between the rotational and translational components in $_f\boldsymbol{\hat{Y}}_1$ and $_f\boldsymbol{\hat{Y}}_2$.
	%The \emph{geometric scale} $L$ has units [length].
	The measure $d$ has units [length] per unit of progress $s$ and quantifies the dissimilarity between the two trajectory shapes as a scalar value in a bi-invariant way. We refer to $d$ as the \textit{Bi-Invariant Local Trajectory-shape Similarity (BILTS)} measure.
	
	\subsection{Properties of the Continuous-time BILTS measure}
	\label{sec:properties}
	
	The BILTS measure quantifies the similarity between two trajectory shapes as one singular scalar value. This measure is non-negative, symmetric, bi-invariant and always bounded. Furthermore, this measure is zero if and only if the compared trajectories have identical bi-invariant shape up to third order. In this subsection, we derive these properties in more detail:
	
	\subsubsection{Symmetry}
	\label{sec:symmetry}
	The measure $d$ is symmetric, since the Frobenius norm of a matrix difference is symmetric. Hence:
	\begin{equation}	
		\left\| _f\boldsymbol{\hat{Y}}_1 - {_f}\boldsymbol{\hat{Y}}_2 \right\|_W = \left\| _f\boldsymbol{\hat{Y}}_2 - {_f}\boldsymbol{\hat{Y}}_1 \right\|_W.
	\end{equation}
	
	\subsubsection{Boundedness}
	\label{sec:bounded}
	We will explain the boundedness of $d$ as follows. The canonical matrix $\boldsymbol{R}(s)$ is essentially the twist $\tw$, acceleration twist $\tw'$, and jerk twist $\tw''$ viewed from another frame (i.e. the functional bi-invariant frame $\{f\}$). Hence, if $\tw$, $\tw'$, and $\tw''$ are bounded, then the elements $r_{ij}$ of $\boldsymbol{R}$ are also bounded. The descriptor ${_f}\boldsymbol{\hat{Y}}(s,\delta s) = \boldsymbol{R}(s)\boldsymbol{C}(s)$ then inherits this boundedness, and so does the similarity measure $d$.
	
	\subsubsection{Bi-invariance}
	
	%The matrix $R(s)$ is both left- and right-invariant in the regular case.
	The BILTS measure $d$ is bi-invariant, meaning that $d$ remains unaffected by variations in both the world and body reference frames across motions. As a result, reference frame variations can no longer disturb motion similarity measurements. 
	%The advantage of this bi-invariance in practice is demonstrated through motion recognition experiments in Section~\ref{sec:recognition}.
	
	To prove that $d$ is bi-invariant, it is sufficient to prove that ${_f}\boldsymbol{\hat{Y}}(s,\delta s)$ is bi-invariant. As already mentioned, ${_f}\boldsymbol{\hat{Y}}(s,\delta s)$ inherits bi-invariance from the bi-invariant matrix $\boldsymbol{R}(s)$. The elements in $\boldsymbol{R}(s)$ are bi-invariant, since they represent the spatial screw twist ${_w}\tw$ and its derivatives expressed in the bi-invariant frame $\{f\}$. In this paper, we also analytically reconfirm the bi-invariance of $\boldsymbol{R}(s)$ by writing the elements of $\boldsymbol{R}(s)$ in function of the bi-invariant ISA-invariants $(\omega_1,\omega_2,\omega_3,v_1,v_2,v_3)$ and their derivatives (see Subsection~\ref{sec:relation}). Bi-invariance of the ISA invariants was already proven in~\cite{DeSchutterJoris2010}. 
	
	\subsubsection{Third-order shape identity}
	The measure $d$ introduced in \eqref{eq:localmetric0} is zero only if $_f\boldsymbol{\hat{Y}}_1(s,\delta s) -{_f}\boldsymbol{\hat{Y}}_2(s,\delta s) = \boldsymbol{0}_{6\times3}$, and this by definition of the Frobenius norm. Applying \eqref{eq:def_Y} to this expression leads to:
	\begin{equation}
		\boldsymbol{R}_1(s)\boldsymbol{C}(\delta s) - \boldsymbol{R}_2(s)\boldsymbol{C}(\delta s) = \boldsymbol{0}_{6\times3}.
	\end{equation}
	Since $\boldsymbol{C}(\delta s)$ is an invertible matrix, we can remove it by post-multiplication with its inverse $\boldsymbol{C}^{-1}(\delta s)$, resulting in:
	\begin{equation}
		\boldsymbol{R}_1(s) - \boldsymbol{R}_2(s) = \boldsymbol{0}_{6\times3}.
	\end{equation}
	Hence, the measure $d$ is zero if and only if $\boldsymbol{R}_1(s) = \boldsymbol{R}_2(s)$. Consequently, $d$ is equal to zero only if the trajectory shapes share the same twist $\tw$ and twist derivatives, $\tw'$ and $\tw''$, when viewed from the bi-invariant frame $\{f\}$. Hence, we conclude that $d$ is zero only if the compared trajectories have identical bi-invariant shape up to third order.
	
	\subsection{Relations with other invariant similarity measures}
	\label{sec:relation}
	Relations exist between the BILTS measure and other similarity measures based on existing invariant descriptors, including the ISA descriptor~\cite{DeSchutterJoris2010} for rigid-body trajectories, the FS descriptor for translation trajectories, and the eFS descriptor~\cite{Vochten2015} and DHB descriptor~\cite{Lee2018} for rigid-body trajectories.
	All these invariant descriptors can be derived from the following generalized model that splits up the motion of the body into two screw twists $\invarobj$ and $\invarmoving$. The twist $\invarobj$ describes the first-order kinematics of the moving body. The twist $\invarmoving$ describes the first-order kinematics of the functional frame upon which the invariants are based. Both twists are expressed in this functional frame \cite{vochten2023invariant}:
	\begin{align} 
		\label{eq:movingframe_and_obj}
		\tf{w}{}'  = \tf{w}{} \skewsym{\invarmoving} ~~~\text{and}~~~
		{_w\tw} = \boldsymbol{S}(\tf{w}{} ) \ {\invarobj}.       
	\end{align}
	
	\subsubsection{Relation with ISA invariants for rigid-body trajectories}
	Following the same convention for the axes of the functional frame as explained in Subsection \ref{sec:theo_BILTS}, the six ISA invariants~\cite{DeSchutterJoris2010} $\left(\omega_1, \omega_2,\omega_3,v_1,v_2, v_3\right)$ relate to the twists $\invarmoving$ and $\invarobj$:
	\begin{align}
		\invarmoving & = \begin{bmatrix}
			\omega_3 & 0 & \omega_2 & v_3 & 0 & v_2
		\end{bmatrix}^T, \label{eq:invarmoving} \\
		\invarobj    & = \begin{bmatrix}
			\omega_1 & 0 & \hspace{2pt}0 & \hspace{4pt}v_1 & 0 & ~0
		\end{bmatrix}^T. \label{eq:invarobj}
	\end{align}
	Remark that this definition of $\invarobj$ is equivalent to the definition of the first column of $\boldsymbol{R}$ in \eqref{eq:matrix_decomp2}. Next, we will establish the relation between the ISA invariants and the other columns of $\boldsymbol{R}$.  First, we obtain the derivative of the twist in \eqref{eq:movingframe_and_obj} by using \eqref{eq:deriv_screw_transform} and factoring out $\boldsymbol{S}(\tf{w}{f})$:
	\begin{equation}
		{_w\tw}' = \boldsymbol{S}(\tf{w}{f}) \left( \skewskewsym{\invarmoving}\invarobj + \invarobj'   \right).
		\label{eq:objd}
	\end{equation}
	Using the same procedure on \eqref{eq:objd} results in:
	\begin{equation}
			{_w\tw}'' = 	\boldsymbol{S}(\tf{w}{f})
			\left( \skewskewsym{\invarmoving}^2\invarobj
			+2 \skewskewsym{\invarmoving} \invarobj'
			+ \skewskewsym{\invarmoving'} \invarobj
			+\invarobj''  \right).
		\label{eq:objdd}
	\end{equation}
	Equations \eqref{eq:movingframe_and_obj}, \eqref{eq:objd} and \eqref{eq:objdd} can be expanded using \eqref{eq:invarmoving} and \eqref{eq:invarobj}. This results in $\boldsymbol{R} =	\boldsymbol{S}^{-1}(\tf{w}{f})\begin{bmatrix}
		{_w\tw} & \hspace{-3pt}{_w\tw'} & \hspace{-3pt}{_w\tw''}
	\end{bmatrix}$
	being equal to:
	\begin{equation}
		\small
			\hspace{-5pt} \boldsymbol{R} =
			\left[\begin{matrix}
				\omega_{1}           & \omega'_{1}                                                         & - \omega_{1} \omega_{2}^{2} + \omega''_{1}                                                                                                      \\
				0                    & \omega_{1} \omega_{2}                                             & \omega_{1} \omega_{2}' + 2 \omega_1' \omega_2                                                                                              \\
				0                    & 0                                                                   & \omega_{1} \omega_{2} \omega_{3}                                                                                                                \\
				v_{1} & v'_{1}                                               & - 2 \omega_1 \omega_2 v_{2} - v_1 \omega_{2}^{2}  +  v''_{1}                                  \\
				0                    &  \omega_1 v_2 +  v_1\omega_2 &  \omega_1 v'_2 + v_1 \omega_{2}' + 2  v'_1 \omega_2 +  2 \omega_{1}' v_2  \\
				0                    & 0                                                                   & v_{1} \omega_{2} \omega_{3} + \omega_{1} v_{2} \omega_3 + \omega_{1} \omega_{2} v_{3}
			\end{matrix}\right].
		\label{eq:qr-isa}
	\end{equation}
	
	This result reconfirms that the screw transformation $S(\tf{w}{f})$ transforms $\left[{_w\tw}~{_w\tw'}~{_w\tw''}\right]$
	into a twice upper-triangular form. This result also shows that $\boldsymbol{R}$ can be explicitly written in function of the ISA-invariants and their derivatives, reconfirming the bi-invariance of $\boldsymbol{R}$. Relation \eqref{eq:qr-isa} also shows that the ISA-invariants are only a partial description of the third-order trajectory shape. 
	The derivatives $\omega_1',~\omega_1'',~\omega_2',~v_1',~v_1''$ and $v_2'$ are additionally required to obtain a complete third-order trajectory-shape description.
	Relation \eqref{eq:qr-isa} can also be used to write the ISA invariants in terms of the elements $r_{ij}$ of $\boldsymbol{R}$:
	\begin{align}
		\omega_1 &= r_{11}                                           & 
		v_1 &=  r_{41}                                               \label {eq:qr-isa-inv-1} \\[3pt]
		\omega_2 &=  \frac{ r_{22} }{ r_{11} }                        & 
		v_2 &= \frac{r_{52}}{r_{11}} - \frac{r_{22} r_{41}}{r_{11}^2} \label {eq:qr-isa-inv-2}\\[2pt]
		\omega_3 &= \frac{ r_{33} }{ r_{22} }                        & 
		v_3 &= \frac{r_{63}}{r_{22}} - \frac{r_{33} r_{52}}{r_{22}^2} \label {eq:qr-isa-inv-3}  
	\end{align}
	
	As explained in Subsection~\ref{sec:properties}, the elements $r_{ij}$ of $\boldsymbol{R}$ are always bounded. However, the same cannot be said of the ISA invariants. From \eqref{eq:qr-isa-inv-1}-\eqref{eq:qr-isa-inv-3}, it can be seen that the invariants $\omega_2$, $v_2$, $\omega_3$, and $v_3$ can become very large when $r_{11}$ and/or $r_{22}$ are small.
	This explains why the ISA descriptor shows a high sensitivity to noise when $r_{11}$ or $r_{22}$ approaches zero. This happens for motions that approximate pure translations ($r_{11} \approx 0$), or rotations about a fixed rotation axis (\mbox{$r_{22} \approx 0$}).
	
	Consequently, defining a similarity measure based on direct comparison of the ISA-invariants has two major drawbacks. First, such a measure would not be bounded, since the ISA-invariants are not bounded. Secondly, the property `third-order shape identity' would not be satisfied, since the ISA-invariants provide only a partial third-order trajectory-shape description.

	\subsubsection{Relation with FS invariants for translation trajectories}
	\label{app:fs_point}
	For FS invariants for translation trajectories, the twists $\invarmoving$ and $\invarobj$ are defined by:
	\begin{align}
		\invarmoving & = \begin{bmatrix}
			\omega_3 & 0 & \omega_2 & v & 0 & 0
		\end{bmatrix}^T, \\
		\invarobj    & = \begin{bmatrix}
			0 & \hspace{6pt}0 & \hspace{2pt}0 & \hspace{4pt}v & 0 & 0
		\end{bmatrix}^T,
	\end{align}
	with $v$ the magnitude of the velocity of the origin of the body frame $\{b\}$ attached to the translating body. 
	The Frenet-Serret invariants for translational trajectories are related to a special form for $\boldsymbol{R}$. We show this by substituting $\invarmoving$ and $\invarobj$ in equations \eqref{eq:movingframe_and_obj}, \eqref{eq:objd}, and \eqref{eq:objdd}. This
	results in ${_w}\boldsymbol{X}$ being equal to:
	\begin{equation}
			\small {_w}\boldsymbol{X}=\boldsymbol{S}( \tf{w}{FS})
			\left[\begin{matrix}0 & 0 & 0\\0 & 0 & 0\\0 & 0 & 0\\
				v & v' & -  v\omega_{2}^{2} + v''\\
				0 &   v\omega_2  &  2 v' \omega_{2}  + v \omega'_{2}\\
				0 & 0 & \omega_{2} \omega_{3} v
			\end{matrix}\right].
		\label{eq:qr-fs}
	\end{equation}
	The lower three rows of the matrix in \eqref{eq:qr-fs} are upper triangular, and identical to the ones in \eqref{eq:qr-isa} in the case of pure translational rigid-body motion ($\omega_1 = \omega_1' = 0$). Hence, for the special case of pure translational rigid-body motion, one solution for the functional frame $\{f\}$ that results in a twice upper triangular form for $\boldsymbol{R}$ is the following: (1) place the origin of $\{f\}$ at the origin of the body frame $\{b\}$, (2) align the orientation of $\{f\}$ with the orientation of the FS-frame for the translation trajectory of the origin of $\{b\}$. 
	
	However, the matrix $\boldsymbol{R}$ obtained from the decomposition \eqref{eq:matrix_decomp} does not naturally obtain the form in \eqref{eq:qr-fs} when describing pure translational motion, since the decomposition \eqref{eq:matrix_decomp} is not unique in that case. For example, another solution for $\{f\}$ is to position it at an `infinite distance' from the moving body, since a pure translation can also be interpreted as infinitesimally small rotation with an infinitely long leverage arm. Hence, to obtain the form of \eqref{eq:qr-fs}, additional regularization is required. 
	
	\subsubsection{Relation with FS invariants for orientation trajectories}
	\label{app:fs_orient}
	For FS invariants for orientation trajectories, the twists $\invarmoving$ and $\invarobj$ are defined by:
	\begin{align}
		\invarmoving & = \begin{bmatrix}
			\omega_3 & 0 & \omega_2 & 0 & 0 & 0
		\end{bmatrix}^T, \\
		\invarobj    & = \begin{bmatrix}
			\omega & ~0 & ~0 & 0 & ~0 & ~0
		\end{bmatrix}^T,
	\end{align}
	with $\omega$ the magnitude of the rotational velocity of the rotating body. The FS invariants for orientation trajectories have a strong relation with the elements in the first three rows of $\boldsymbol{R}$. We show this by substituting $\invarmoving$ and $\invarobj$ in equations \eqref{eq:movingframe_and_obj}, \eqref{eq:objd} and \eqref{eq:objdd}. This
	results in ${_w}\boldsymbol{X}$ being equal to:
	\begin{equation}
			\small {_w}\boldsymbol{X}=\boldsymbol{S}(\tf{w}{FS})
			\left[\begin{matrix}\omega_{1} & \omega'_{1} & - \omega_{1} \omega_{2}^{2} + \omega''_{1} \\
				0 & \omega_{1} \omega_{2} & \omega_1\omega'_2 +2 \omega'_1\omega_2 \\0 & 0 & \omega_{1} \omega_{2} \omega_{3}\\
				0          & 0           & 0                                          \\
				0          & 0           & 0                                          \\
				0          & 0           & 0
			\end{matrix}\right].
		\label{eq:qr-efs}
	\end{equation}
	The upper three rows of the matrix in \eqref{eq:qr-efs} are identical to the ones in \eqref{eq:qr-isa}. Hence, the orientation of $\{f\}$ for rigid-body trajectories always coincides with the orientation of the FS frame when applied to the body's orientation trajectory.

	\section{Discretized BILTS and numerical computation}
	\label{sec:practical}
	In Section \ref{sec:descriptor}, we introduced the BILTS measure $d$, which is based on the bi-invariant trajectory-shape descriptor ${_f}\boldsymbol{\hat{Y}}(s,\delta s)$. The derivation of ${_f}\boldsymbol{\hat{Y}}(s,\delta s)$ was based on a continuous-time assumption for the rigid-body trajectory $\tf{w}{b}(s)$. Furthermore, it was assumed that the twist ${_w}\tw(s)$ and its derivatives ${_w}\tw'(s)$ and ${_w}\tw''(s)$ were known. 
	
	In this section, we first explain the numerical computation of the twice upper-triangular matrix $\boldsymbol{R}(s)$ by introducing an extended QR-decomposition algorithm. Afterwards, we introduce a discretized bi-invariant descriptor that approximates the continuous-time descriptor ${_f}\boldsymbol{\hat{Y}}(s,\delta s)$. We then show how this discretized descriptor can be calculated from a discrete-time rigid-body trajectory without requiring the explicit estimation of the higher order twist derivatives ${_w}\tw'(s)$ and ${_w}\tw''(s)$. Lastly, we introduce regularization to increase the BILTS measure's robustness near singularities in the trajectory.
	\subsection{Calculation of R using an extended QR-decomposition}
	\label{sec:eQR}
	This section explains an \textit{extended QR-decomposition (eQR)} to calculate the twice upper-triangular matrix $\boldsymbol{R}$ and functional frame~$\{f\}$ from the matrix ${_w}\boldsymbol{X}(s)$. 
	
	The decomposition of ${_w}\boldsymbol{X}(s)$ in \eqref{eq:matrix_decomp} can be expanded using the definition of the screw transformation matrix in \eqref{eq:screwtf}:
	\begin{equation}
			{_w}\boldsymbol{X}(s)   =
			\begin{bmatrix}
				\boldsymbol{X}_1 \\
				\boldsymbol{X}_2
			\end{bmatrix}
			= \begin{bmatrix}
				\boldsymbol{Q}             & \boldsymbol{0} \\
				\skewsym{\boldsymbol{p}} \boldsymbol{Q} & \boldsymbol{Q}
			\end{bmatrix}
			\begin{bmatrix}
				\boldsymbol{R}_1 \\
				\boldsymbol{R}_2
			\end{bmatrix} .
		\label{eq:extendedQR}
	\end{equation}
	The first three rows of this equation are:
	\begin{equation}
		\boldsymbol{X}_1 = \boldsymbol{Q} \boldsymbol{R}_1,
		\label{eq:A1}
	\end{equation}
	where the first two columns of $\boldsymbol{Q}$ and $\boldsymbol{R}_1$ can be found using a standard QR-decomposition~\cite{golub2013matrix} assuming the diagonal elements $r_{11}$ and $r_{22}$ differ from zero. Afterwards, we can impose that $r_{11}>0$ and $r_{22}>0$ by flipping the sign of the corresponding row of $\boldsymbol{R}_1$ and the corresponding column of $\boldsymbol{Q}$ whenever necessary.
	The third column of $\boldsymbol{Q}$ and the sign of the third diagonal element $r_{33}$ is then determined by imposing that $\boldsymbol{Q}$ is a right-handed orthogonal matrix.  The resulting $\boldsymbol{Q}$ is then unique. Note that we do not require $\boldsymbol{X}_1$ to be invertible, since $r_{33}$ is still allowed to be zero.
	
	Thus far, we have completely and uniquely determined the matrices $\boldsymbol{Q}$ and $\boldsymbol{R}_1$. The position vector $\boldsymbol{p}$ in \eqref{eq:extendedQR} can then be determined as follows. After pre-multiplication with $\boldsymbol{Q}^T$, the last three rows of \eqref{eq:extendedQR} can be written as:
	\begin{align}
		\boldsymbol{Q}^T \boldsymbol{X}_2 & = ( \boldsymbol{Q}^T \skewsym{\boldsymbol{p}} \boldsymbol{Q} ) \boldsymbol{R}_1 + \boldsymbol{R}_2, \\
		& = ( \skewsym{\boldsymbol{Q}^T \boldsymbol{p} \hspace{2pt} } ) \boldsymbol{R}_1 + \boldsymbol{R}_2,          \\
		& = \skewsym{\boldsymbol{p}^{*}} \boldsymbol{R}_1 + \boldsymbol{R}_2.
		\label{eq:extendqr-origin}
	\end{align}
	where $\boldsymbol{p}^{*} = \boldsymbol{Q}^T \boldsymbol{p}$ represents the position vector from the origin of the spatial frame $\{w\}$ to the origin of the functional frame $\{f\}$, with coordinates $(x~y~z)^T$ expressed in $\{f\}$. From the lower-diagonal elements in \eqref{eq:extendqr-origin} and by imposing an upper-triangular matrix $\boldsymbol{R}_2$, three scalar equations are obtained that do not rely on the unknown upper-triangular elements in $\boldsymbol{R}_2$:
	\begin{align}
		(\boldsymbol{Q}^T \boldsymbol{X}_2)_{21} &= r_{11} z, \\
		(\boldsymbol{Q}^T \boldsymbol{X}_2)_{31} &= -r_{11} y, \\          
		(\boldsymbol{Q}^T \boldsymbol{X}_2)_{32} &= r_{22} x - r_{12} y, 	\label{eq:extendedqr_scalar}
	\end{align}
	out of which the coordinates  $x,~y,~z$ of $\boldsymbol{p}^*$ can be uniquely determined again assuming $r_{11}>0$ and $r_{22}>0$.
	Now that $\boldsymbol{Q}$, $\boldsymbol{R}_1$, and $\boldsymbol{p}^*$ are completely determined, $\boldsymbol{p}$ can be determined by $\boldsymbol{p} = \boldsymbol{Q} \boldsymbol{p}^*$, while $\boldsymbol{R}_2$ can be solved from~\eqref{eq:extendqr-origin}:
	\begin{align}
		\boldsymbol{R}_2 & = \boldsymbol{Q}^T \boldsymbol{X}_2 - \skewsym{\boldsymbol{p}^{*}} \boldsymbol{R}_1 \label{eq:R2-computation}.
	\end{align}
	The twice upper-triangular matrix $\boldsymbol{R} = \small \Big[\begin{array}{c} \boldsymbol{R}_1  \\ \boldsymbol{R}_2 \end{array}\Big]$ together with the orientation $\boldsymbol{Q}$ and position $\boldsymbol{p}$ of the functional frame $\{f\}$ are now \textit{completely and uniquely} determined.
	
	In the \textit{singular} case where $r_{11}=0$ or $r_{22}=0$, this eQR-decomposition is degenerate. In that case, additional assumptions are required to obtain a unique solution for $\boldsymbol{R}$, $\boldsymbol{Q}$, and $\boldsymbol{p}$. Section~\ref{sec:robust} proposes a robust approach to deal with these singular cases.
	
	\subsection{Discretized bi-invariant trajectory-shape descriptor}
	\label{sec:discrete}
	In practice, it is reasonable to assume that the trajectory is available as an equidistantly sampled sequence of poses $\tf{w}{b}(s_k)$ with $s_k = k\Delta s$, $\Delta s$ the step size, and $k$ ranging from zero to the total number of samples $N$. In this subsection, we introduce the discretized descriptor ${_f}\boldsymbol{Y}(s_k,\Delta s)$ which approximates the continuous-time descriptor ${_f}\boldsymbol{\hat{Y}}(s=s_k,\delta s = \Delta s)$, as visualized in Figure~\ref{fig:shapes_discrete}:
	\begin{figure}[t]
		\centering
		\includegraphics[width=0.75\linewidth]{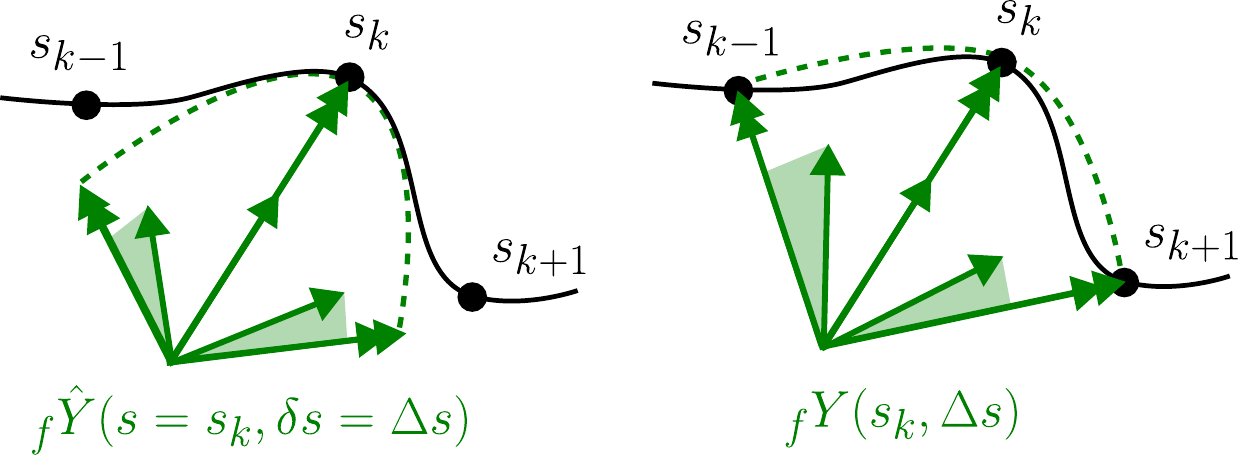}
		\caption{Visualization of the discretized descriptor ${_f}\boldsymbol{Y}(s_k,\Delta s)$ (right figure) as a discrete approximation of the continous-time descriptor ${_f}\boldsymbol{\hat{Y}}(s = s_k, \delta s = \Delta s)$ (left figure). }
		\vspace{-5pt}
		\label{fig:shapes_discrete}
	\end{figure}
	
	The discretized descriptor can be directly calculated from a sequence of spatial screw twists.  The spatial screw twist at $s_k$ can be computed from trajectory samples $\tf{w}{b}(s_{k-1})$ and $\tf{w}{b}(s_{k+1})$ using the \textit{matrix logarithm} \cite{lynch2017modern} on $SE(3)$:
	\begin{equation}
		\skewsym{{_w\tw}(s_k)} = \frac{1}{2\Delta s} \log \left( \tf{w}{b}(s_{k+1}) \ \tf{w}{b}^{-1} (s_{k-1})\right).
		\label{eq:logm}
	\end{equation}
	To explain the computation of the discretized bi-invariant descriptor ${_f}\boldsymbol{Y}(s_k,\Delta s)$, we again first introduce the discretized right-invariant descriptor ${_w}\boldsymbol{Y}(s_k,\Delta s)$. It simply consists of the spatial screw twists at the instances $s_{k-1}$, $s_k$, and $s_{k+1}$:
	\begin{equation}
			\small {_w}\boldsymbol{Y}(s_k,\Delta s) = \begin{bmatrix} {_w}\tw(s_{k-1}) & {_w}\tw(s_k) & {_w}\tw(s_{k+1}) \end{bmatrix}.
	\end{equation}
	The discretized descriptor ${_w}\boldsymbol{Y}(s_k,\Delta s)$ approximates the continuous-time descriptor ${_w}\boldsymbol{\hat{Y}}(s=s_k,\delta s=\Delta s)$ since:
	\begin{equation}
		\small {_w}\boldsymbol{Y}(s_k, \Delta s) - {_w}\boldsymbol{\hat{Y}}(s = s_k,\delta s = \Delta s) = \boldsymbol{0}_{6\times3} + \mathcal{O}(\Delta s^3).
	\end{equation}
	The truncation error $\mathcal{O}(\Delta s^3)$ is due to the second-order Taylor series expansions of the twist ${_w}\tw(s)$ in ${_w}\boldsymbol{\hat{Y}}(s,\delta s)$. 
	%Note that we denote the discretized descriptors ${_w}Y(s_k, \Delta s)$ and ${_f}Y(s_k, \Delta s)$ without the hat $\hat{~}$~, since they do not involve these second-order Taylor-series expansions.
	
	The discretized bi-invariant descriptor ${_f}\boldsymbol{Y}(s_k, \Delta s)$ is then obtained by transforming the spatial frame $\{w\}$ of the discretized descriptor ${_w}\boldsymbol{Y}(s_k, \Delta s)$ to a functional frame:
	\begin{equation}
		\label{eq:trans_discrete}
		{_f}\boldsymbol{Y}(s_k, \Delta s) = \boldsymbol{S}^{-1}(\tf{w}{f}){_w}\boldsymbol{Y}(s_k, \Delta s),
	\end{equation}
	such that the discretized descriptor ${_f}\boldsymbol{Y}(s_k, \Delta s)$ has the same structure as the continuous-time descriptor ${_f}\boldsymbol{\hat{Y}}(s, \delta s)$ in~\eqref{eq:BILTS_B_matrix}:
	\begin{equation*}
		\small \label{eq:Y_can2}
		_f\boldsymbol{Y}(s_k,\Delta s) = 
			\begin{bmatrix}
				\omega_x^- & \omega_x & \omega_x^+ \\
				\omega_y^- & 0 & \omega_y^+ \\
				\omega_z^{-} & 0 & \omega_z^+\\
				v_x^- & v_x & v_x^+ \\
				v_y^- & 0 & v_y^+ \\
				v_z^- & 0 & v_z^+
			\end{bmatrix}
			, ~\text{with} ~ \left\{ \begin{array}{c} \hspace{17pt}\omega_x \geq 0 \\ 
				\hspace{-10pt}\omega_y^+-\omega_y^- \geq 0 \\ 
				\hspace{23pt}\omega_z^- = \omega_z^+ \\ 
				\hspace{23pt}v_z^- = v_z^+
			\end{array} \right. \hspace{-5pt}\tag{r.\ref{eq:BILTS_B_matrix}}
	\end{equation*} 
	
	This can be done using the same eQR-decomposition algorithm as introduced in Sec.~\ref{sec:eQR}. To explain this, we show that $_f\boldsymbol{Y}(s_k,\Delta s)$ can be transformed to a twice upper-triangular matrix by right-multiplication with an invertible matrix $\boldsymbol{A}$:
	\begin{equation}
		\label{eq:Y_can3}
		\small 
			_f\boldsymbol{Y}(s_k,\Delta s)\boldsymbol{A} = 
			\begin{bmatrix}
				\omega_x & \omega_x^+ - \omega_x^- &  \omega_x^{-}\\
				0 & \omega_y^+ - \omega_y^- & \omega_y^{-} \\
				0 & 0 & \omega_z^- \\
				v_x &  v_x^+ - v_x^- &  v_x^-\\
				0 & v_y^+ - v_y^- & v_y^-  \\
				0 & 0 & v_z^- 
			\end{bmatrix}, \text{with}~\boldsymbol{A} = \begin{bmatrix} 0 & -1 & 1 \\
				1 & 0 & 0 \\
				0 & 1 & 0
		\end{bmatrix}.
	\end{equation}
	Remark that the second column of $_f\boldsymbol{Y}(s_k,\Delta s) \boldsymbol{A}$ in \eqref{eq:Y_can3} is closely related to a central finite differences scheme to estimate the twist derivative at $s_k$. Pre- and post-multiplication of \eqref{eq:trans_discrete} with $\boldsymbol{S}(\tf{w}{f})$ and $\boldsymbol{A}$, respectively, results in:
	\begin{equation}
		\label{eq:eQR_discrete}
		_w\boldsymbol{Y}(s_k,\Delta s)\boldsymbol{A} = \boldsymbol{S}(\tf{w}{f}) \ {_f}\boldsymbol{Y}(s_k,\Delta s)\boldsymbol{A},
	\end{equation} 
	Hence, 	\eqref{eq:eQR_discrete} shows that $_w\boldsymbol{Y}(s_k,\Delta s)\boldsymbol{A}$ can be decomposed into a screw transformation matrix and a twice upper-triangular matrix ${_f}\boldsymbol{Y}(s_k,\Delta s)\boldsymbol{A}$. Consequently, we can find ${_f}\boldsymbol{Y}(s_k,\Delta s)$ by first calculating the eQR-decomposition of $_w\boldsymbol{Y}(s_k,\Delta s)\boldsymbol{A}$. Afterwards, we post-multiply the found twice upper-triangular matrix ${_f}\boldsymbol{Y}(s_k,\Delta s)\boldsymbol{A}$ with $\boldsymbol{A}^{-1}$ to obtain ${_f}\boldsymbol{Y}(s_k,\Delta s)$. 
	
	Above eQR-decomposition is regular when the diagonal elements $\omega_x$ and \mbox{$\omega_y^+ - \omega_y^-$} in \eqref{eq:Y_can3} differ from zero. This regularity condition for the discretized descriptor ${_f}\boldsymbol{Y(}s_k,\Delta s)$ is closely related to the regularity condition ($r_{11}\neq 0$ and $r_{22}\neq 0$) for the continuous-time descriptor ${_f}\boldsymbol{\hat{Y}}(s,\delta s)$.
	
	\subsection{Progress scale of the discretized descriptor}
	In Section~\ref{sec:descriptor}, the parameter $\delta s$, controlling the sensitivity of the numbers in $_w\boldsymbol{\hat{Y}}(s,\delta s)$ to variations in the twist derivatives was free to choose and could take any continuous value. In the definition of the discretized descriptor ${_w}\boldsymbol{Y}(s,\Delta s)$, this degree of freedom is lost. The $\Delta s$ in ${_w}\boldsymbol{Y}(s,\Delta s)$ corresponds to the discretization resolution of the trajectory $\tf{w}{b}(s_k)$. In practice, $\Delta s$ cannot be chosen freely, but must be sufficiently small to preserve high-frequency components within the trajectory. 
	
\begin{figure}[t]
		\centering
		\includegraphics[width=0.6\linewidth]{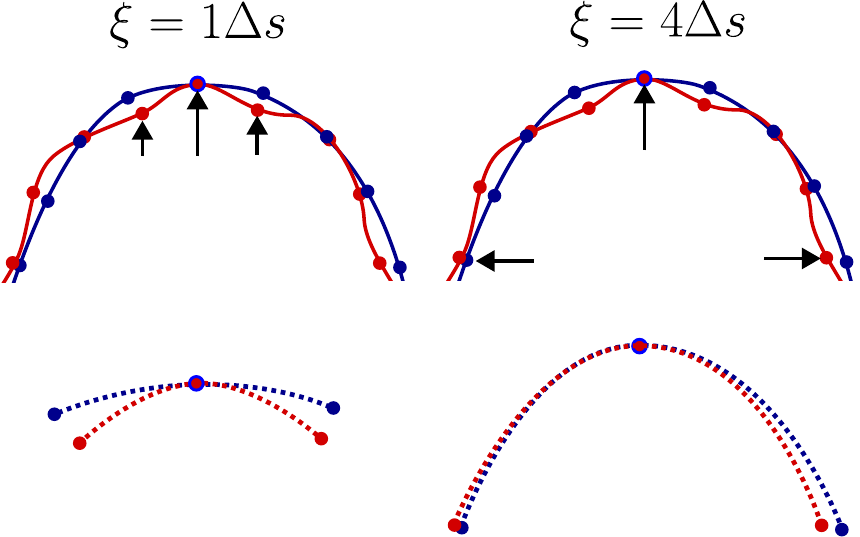}
		\caption{Conceptual illustration of the importance of specifying the progress scale $\xi$ when measuring trajectory-shape similarity. The shapes of two point trajectories (in red and blue) are shown at two progress scales. At the finer scale $\xi = 1 \Delta s$ (left), the shapes (bottom left) show higher dissimilarity due to small local variations. In contrast, at the larger scale $\xi = 4 \Delta s$ (right), this dissimilarity is reduced (bottom right), revealing higher trajectory shape similarity when evaluated at a larger progress scale.}
		\vspace{-3pt}
		\label{fig:progress_scale}
	\end{figure}

	To reintroduce a similar degree of freedom, we introduce the integer multiple $m$ of the discretization resolution $\Delta s$ as the discretized \textit{progress scale} $m\Delta s$ of the trajectory-shape descriptor. This progress scale represents a discretization of a desired continuous value for the progress scale $\xi$ as follows: 
	\begin{equation}
		m\Delta s \approx \xi ~~~\text{with}~~~ m = \text{max}\left(1,\text{round}(\xi / \Delta s)\right)
	\end{equation}
	Specifying the progress scale $\xi$ is essential when comparing trajectory shapes, since shapes that are dissimilar at finer scales due to small local variations may still be similar when viewed at larger scales. This concept is illustrated in Figure~\ref{fig:progress_scale}.
	
\begin{figure}[t]
		\centering
		\includegraphics[width=0.45\linewidth]{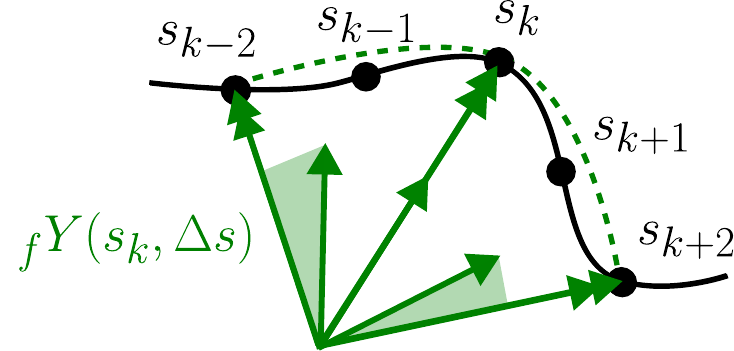}
		\caption{Visualization of the descriptor ${_f}\boldsymbol{Y}(s_k,2\Delta s)$ in the case of a finer progress resolution $\Delta s$ compared to Figure~\ref{fig:shapes_discrete}, but a progress scale of $2\Delta s$.}
		\vspace{-5pt} 
		\label{fig:shapes_discrete2}
	\end{figure}

	We define the right-invariant descriptor ${_w}\boldsymbol{Y}(s_k,m\Delta s)$ with discretized progress scale of $m\Delta s$ as follows, using the twists at the instances $s_k-m\Delta s$ and $s_k+m\Delta s$:
	\begin{equation}
			{_w}\boldsymbol{Y}(s_k,m\Delta s) = \left[{_w}\tw(s_k-m\Delta s) ~~{_w}\tw(s_k) ~~{_w}\tw(s_k+m\Delta s)\right].
	\end{equation} 
	Finally, the discretized bi-invariant descriptor ${_f}\boldsymbol{Y}(s_k,m\Delta s)$ with progress scale of $m\Delta s$ can be computed directly from ${_w}\boldsymbol{Y}(s_k,m\Delta s)$ using the eQR-decomposition explained in Subsection~\ref{sec:discrete}. Figure~\ref{fig:shapes_discrete2} visualizes ${_f}\boldsymbol{Y}(s_k,m\Delta s)$ for $m=2$.
	
	\subsection{Introducing the discretized BILTS measure}
	Similarly to the continuous-time BILTS measure in \eqref{eq:localmetric0}, the discretized BILTS measure $d_k$ is defined as:
	\begin{align}
		d_k ~\defeq~ & 
		\left\| {_f} \boldsymbol{Y}_1(s_k, m\Delta s) - {_f}\boldsymbol{Y}_2(s_k,m\Delta s) \right\|_{\boldsymbol{W}},
		\label{eq:localmetric_discrete}
	\end{align}
	with $\boldsymbol{W} = \text{diag}\left(L^2,L^2,L^2,1,1,1\right)$. This measure quantifies the difference between two discretized trajectory shapes as one scalar value in a bi-invariant manner. It consists of two parameters: $L$ [length] and $m$ [-]. The parameter $L$ determines the weighting between rotations and translations within the rigid-body motion. The value $m\Delta s$ determines the discretized progress scale at which the trajectory shapes are evaluated. 
	\subsection{Enhancing robustness to singularities} 
	\label{sec:robust}
	The eQR-decomposition in Subsection~\ref{sec:eQR} is degenerate when the first or the second diagonal element of the twice upper-triangular matrix is zero. For the continuous-time case, this corresponds to $r_{11}$ or $r_{22}$ being zero. For the discretized case, this corresponds to $\omega_x$ or \mbox{$\omega_y^+-\omega_y^-$} being zero. In these singular cases, the eQR-decomposition is no longer unique. Consequently, the orientation $\boldsymbol{Q}$ and origin $\boldsymbol{p}$ of the functional frame $\{f\}$ will be highly sensitive to measurement noise when approaching these singular cases. Typically, the origin of $\{f\}$ diverges significantly from the location of the moving body, while the orientation of $\{f\}$ is subject to instabilities and frequent flips. This noise sensitivity of $\{f\}$ is then inherited by the corresponding trajectory-shape descriptors and, hence, also by the similarity measures $d$ and $d_k$. In this subsection, we explain how to increase the similarity measure's robustness to noise when approaching these singular cases.
	
	\subsubsection{Regularization of the origin of $\{f\}$}
\begin{figure}[t]
		\centering
		\includegraphics[width=0.6\linewidth]{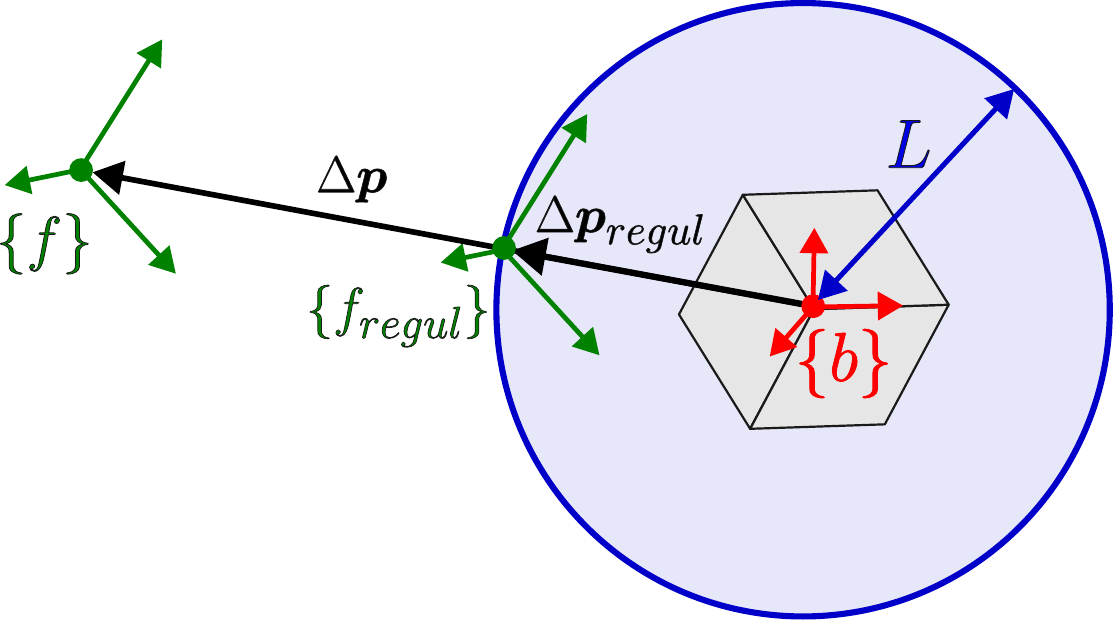}
		\caption{Regularization of the origin of the functional frame $\{f\}$. The distance vector $\Delta \boldsymbol{p}$ between the origin of the body frame $\{b\}$ and functional frame $\{f\}$ is limited to a predefined maximum magnitude of $L$. This results in the regularized distance vector $\Delta p_{regul}$ and functional frame $\{f_{regul}\}$.}
		\vspace{-10pt}
		\label{fig:regularization}
	\end{figure}

	First, we regularize the origin of $\{f\}$ in a manner inspired by the regularization strategy in \cite{verduyn2023enhancingmotiontrajectorysegmentation}. In~\cite{verduyn2023enhancingmotiontrajectorysegmentation}, the authors showed that limiting the ISA to a predefined distance from the body frame resulted in an increased robustness of their trajectory segmentation approach when dealing with singularities. Following a similar strategy, we regularize the origin of $\{f\}$ to remain within a distance $L$ from the origin of $\{b\}$ (see Figure~\ref{fig:regularization}). We propose to use the same parameter $L$ for both this regularization action and the weighting in \eqref{eq:localmetric0} and \eqref{eq:localmetric_discrete}. The rationale behind this choice is discussed at the end of this subsection.
	%This regularization action is explained in the next paragraph.
	
	This regularization action can be implemented as follows. After performing \eqref{eq:A1} and \eqref{eq:extendedqr_scalar} of the eQR-decomposition to obtain $\boldsymbol{Q}$, $\boldsymbol{R}_1$ and $\boldsymbol{p}^*$, we calculate the position vector $\Delta \boldsymbol{p}$ from the origin of $\{b\}$ to the origin of $\{f\}$ by:
	\begin{equation}
		\Delta \boldsymbol{p} = \boldsymbol{p}^* - \boldsymbol{Q}^T\boldsymbol{p}^{b},
	\end{equation}
	with $\boldsymbol{p}^{b}$ the position vector from the origin of $\{w\}$ to the origin of $\{b\}$ with coordinates expressed in $\{w\}$. We then regularize this vector $\Delta \boldsymbol{p}$ to a maximum magnitude $L$ by:
	\begin{equation}
		\Delta \boldsymbol{p}_{regul} = \left\{ \begin{array}{cc} \Delta \boldsymbol{p} &\text{if}~~\left\Vert \Delta \boldsymbol{p} \right\Vert \leq L \\
			L \displaystyle\frac{\Delta \boldsymbol{p} }{\left\Vert\Delta \boldsymbol{p} \right\Vert} &\text{if}~~\left\Vert\Delta \boldsymbol{p} \right\Vert > L 
		\end{array} \right.
	\end{equation}
	The \hspace{-1pt}position of the regularized origin of $\{f\}$ is then found by:
	\begin{equation}
		\boldsymbol{p}^*_{regul} =  \boldsymbol{Q}^T\boldsymbol{p}^{b} + \Delta \boldsymbol{p}_{regul}, 
	\end{equation}
	and the regularized matrix $\boldsymbol{R}_{2,regul}$ is then found using~\eqref{eq:R2-computation}:
	\begin{align} 
		\boldsymbol{R}_{2,regul} & = \boldsymbol{Q}^T \boldsymbol{X}_2 - \skewsym{\boldsymbol{p}^*_{regul}} \boldsymbol{R}_1. \hspace{-5pt}\tag{r.\ref{eq:R2-computation}}
	\end{align}
	When above regularization is active ($\Vert \Delta \boldsymbol{p} \Vert > L$), the matrix $\boldsymbol{R}_{2,regul}$ will no longer be strictly upper-triangular, and the values in $\boldsymbol{R}_{2,regul}$ will depend on the location of the origin of $\{b\}$. Hence, only when near singularities, we sacrifice invariance to the origin of $\{b\}$ to increase the eQR-decomposition's robustness to noise. 
	%Nevertheless, the sensitivity of the values in $\boldsymbol{R}_{2,regul}$ to variations in the origin of $\{b\}$ typically remains limited when approaching these singular cases, especially in the case of pure translations~\cite{verduyn2023enhancingmotiontrajectorysegmentation}. 
	
	Because we use the same parameter $L$ for both the regularization described above, and the weighting in \eqref{eq:localmetric0} and \eqref{eq:localmetric_discrete}, the similarity measures $d$ and $d_k$ for rigid-body trajectories naturally reduce to similarity measures for point trajectories when setting $L=0$. That is, for $L=0$, the origin of $\{f\}$ will be forced to coincide with the origin of $\{b\}$, and $\boldsymbol{R}_2$ will consequently describe the kinematics of the origin of $\{b\}$. Furthermore, the similarity measures $d$ and $d_k$ will only focus on the shape of the translation trajectory of the origin of $\{b\}$, since the shape of the rotation trajectory of $\{b\}$ will be neglected due to the weighting with $L=0$ in \eqref{eq:localmetric0} and \eqref{eq:localmetric_discrete}. Consequently, this case of $L=0$ results in a trajectory-shape similarity measure that is strongly related to the FS descriptor approach for pure translations, as explained in Sec.~\ref{app:fs_point}.
	
	\subsubsection{Introducing a post-processing orientation alignment}
	In \cite{DSRF2018} and \cite{verduyn2023enhancingmotiontrajectorysegmentation}, it is explained how the orientation of two frames can be spatially aligned based on a singular value decomposition of a weighted covariance matrix between twist coordinates seen from the two frames. Since the descriptor ${_f}\boldsymbol{Y}(s_k,m\Delta s)$ contains twist coordinates, we can use this alignment approach to robustly deal with instability of the orientation of $\{f\}$ when approaching singularities. That is, we can add freedom in the choice for the orientation of $\{f\}$ when evaluating the shape similarity $d_k$ as follows:
	\begin{equation}
		\label{eq:d_regul}
		d_k = 
		\min_{\boldsymbol{P}}\left\| \boldsymbol{P}\ {_f}\boldsymbol{Y}_1(s_k,m\Delta s) - {_f}\boldsymbol{Y}_2(s_k,m\Delta s) \right\|_W,
	\end{equation}
	with $\boldsymbol{P}$ the optimal orientation transformation matrix: 
	\begin{equation}
		\boldsymbol{P} = \text{blkdiag}\left(\boldsymbol{R},\boldsymbol{R}\right) ~~~\text{with}~~~\boldsymbol{R}\in SO(3),
	\end{equation}
	which can be obtained using methods such as the Kabsch algorithm~\cite{kabsch1976solution}. This orientation alignment will result in a higher noise robustness of $d_k$ when approaching singularities. However, this improved robustness is achieved at the expense of an increased computational effort for evaluating $d_k$.
	
	In conclusion, since the descriptor ${_f}\boldsymbol{Y}(s_k,m\Delta s)$ consists of a sequence of twists, it facilitates an orientation alignment as a post-processing operation. This orientation alignment enables a more robust comparison of trajectory shapes near singularities. This is a unique advantage of BILTS compared to measures derived from the ISA, eFS, or DHB invariants, which do not facilitate such an orientation alignment operation.
	
	\section{BILTS validation for trajectory recognition}
	\label{sec:recognition}
	In Section~\ref{sec:descriptor}, we introduced the continuous-time BILTS measure $d$ that quantifies the similarity between trajectory shapes in a bi-invariant manner. In Section~\ref{sec:practical}, we introduced the discretized BILTS measure $d_k$ and explained how to compute it from a discretized rigid-body trajectory $\tf{w}{b}(s_k)$. This section details the application of the BILTS measure $d_k$ for the recognition of object-manipulation tasks in diverse contexts. This is followed by an experimental validation and comparison with other invariant similarity measurement approaches.
	
	\subsection{Recognition Datasets: Challenge of Contextual Variations}
	Our goal is to enable the robust recognition of object-manipulation tasks when performed in diverse contexts. This challenge of dealing with diverse contexts is present in the existing \textit{Daily-life activities} (DLA) dataset and the novel \textit{Synthetic} (SYN) rigid-body motion dataset. Both datasets have been made publicly accessible\footnote{Available on Zenodo at \url{https://doi.org/10.5281/zenodo.12806232} .}.
	
	The DLA dataset consists of ten classes of activities: \textit{cutting, painting, pouring, stowing, quarter-turn, shaking, wiping, sinusoidial motion, scooping,} and \textit{scooping followed by pouring}. The activities were repeatedly executed with respect to three different viewpoints, and, for every viewpoint, the demonstrator was instructed to perform the motion using four different execution styles: (1) normal; (2) with a larger spatial scale; (3) with a different time profile; and (4) with a longer duration. This resulted in a total of $3\times4=12$ different contexts. 
	
	The trials were recorded using a Krypton K600 camera from NIKON Metrology by tracking up to nine LED markers attached to the manipulated object. For each trial, a body frame was extracted from three LED markers that remained visible during motion execution, with their average position serving as the frame's origin. The body frame’s orientation was extracted from the LED markers using a Gram-Schmidt orthonormalization algorithm. However, due to changes in the sensor viewpoint and varying LED occlusions, different body frames were extracted across trials, leading to significant variations in the orientation of the body frame across all trials.
	
	To introduce even more contextual variations, we developed two adapted DLA datasets, \mbox{\textit{adapted DLA 1}} and \mbox{\textit{adapted DLA 2}}. The first adaptation was designed to underscore the drawbacks of approaches that lack right-invariance. For this adaptation, the reference point on the object was displaced by 20cm along the x-axis of the body frame, and this for the motions seen from the first and third viewpoint. The second adaptation was designed to underscore the drawbacks of approaches that lack left-invariance and true locality. For this adaptation, we copied each motion, randomly changed the execution direction of this copy by $\pm90\degree$ along the z-axis of the world frame, and concatenated it with the original motion. 
	%The specific sign, plus or minus $45\degree$, randomly varied across motion classes and contexts. 
	Hence, the \mbox{\textit{adapted DLA 2}} dataset consists of trials of successive motions performed in diverse directions. 
	
	We also developed and released the novel SYN dataset. It consists of seven elementary rigid-body motions: \textit{linear, circular, and helical translations; fixed-axis and precession rotations; and screw motions with positive and negative pitch}. Brown noise was added to each synthetically generated rigid-body trajectory by integrating white noise at the velocity level. This Brown noise simulated lower-frequency human-induced variations in the motions. Additionally, the trajectories were represented in various world and body reference frames. This resulted in three different contexts: \textit{Original}, \textit{Change in References 1}, and \textit{Change in References~2}. 
	
	\subsection{Invariant Similarity: A Solution for Contextual Variations}
	
	To enable the robust recognition of object-manipulation tasks when performed in diverse contexts, we propose to classify motion trajectories based on their similarity to known reference trajectories in a manner that is invariant to contextual variations. The SYN and DLA datasets include both world and body frame variations. The DLA dataset additionally includes time profile variations. To deal with these variations, we propose to first transform the temporal motion trajectories to geometric trajectories. This allows the motions to be classified purely based on spatial geometry, independent of the time profile. We then compute invariant trajectory descriptors from these geometric trajectories to robustly deal with body and world frame variations.
	
	\subsubsection{Geometric Trajectories: Mitigating Time Dependency}
	
	The temporal trajectories $\tf{w}{b}(t_k)$, where $t_k$ represents the time at sample $k$, were transformed to geometric trajectories $\tf{w}{b}(s_i)$, where $s_i$ represents the geometric progress at \mbox{sample~$i$}. To do this numerically, we first determined the traversed geometric progress $s_k$ at each timesample $k$ by: 
	\begin{equation}
		s(t_k) = \sum_{j=1}^{k-1} \dot{s}(t_j) \Delta t~,
		\label{eq:num_progress}
	\end{equation}
	with $\Delta t$ the time interval between timesamples.
	
	Different definitions for the geometric progress $s$ (\mbox{\textit{arclength}}, \textit{angle}, or \textit{screw path}) were used in the experiments. 
	In case of \textit{arclength}, the traversed arclength along the path traced by the reference point on the body was used as the geometric progress: \mbox{$\dot{s}(t_j) = \Vert \boldsymbol{\dot{p}}^b(t_j) \Vert$}.  In case of \textit{angle}, the traversed angle of the moving body was used as the geometric progress \mbox{$\dot{s}(t_j) = \Vert \boldsymbol{\omega}(t_j) \Vert$}. In case of \textit{screw path}, the regulated screw-based progress~\cite{verduyn2023enhancingmotiontrajectorysegmentation}, consisting of both rotational and regulated translational velocity along the ISA was used.
	% : \mbox{$\dot{s}(t_j) = \sqrt{ L^2\Vert\boldsymbol{\omega}(t_j) \Vert^2 + \Vert\boldsymbol{\tilde{v}}^{ISA}(t_j)\Vert^2}$}.
	
	Note that using \textit{arclength} as the geometric progress is not right-invariant. Therefore, we did not utilize this type of geometric progress for the bi-invariant approaches (ISA and BILTS), as doing so would compromise their right-invariance.
	
	The velocities $\boldsymbol{\dot{p}}^b(t_j)$, $\boldsymbol{\omega}(t_j)$, and $\boldsymbol{\tilde{v}}^{ISA}(t_j)$ were estimated from the temporal rigid-body trajectories using numerical first-order finite differences schemes. The velocity of the reference point on the moving body $\boldsymbol{\dot{p}}^b(t_j)$ was found by: $$\boldsymbol{\dot{p}}^b(t_j) = \frac{\boldsymbol{p}^b(t_{j+1})-\boldsymbol{p}^b(t_{j})}{\Delta t}.$$
	The rotational velocity $\boldsymbol{\omega}(t_j)$ at each time sample $t_j$ was obtained from the spatial screw twist ${_w\tw}(t_j)$ which was calculated from subsequent poses $\tf{w}{b}(t_j)$ and $\tf{w}{b}(t_{j+1})$ using the \textit{matrix logarithm} on $SE(3)$, such as in \eqref{eq:logm}:
	\begin{equation*}
		\skewsym{{_w\tw}(t_j)} = \frac{1}{\Delta t} \log \left( \tf{w}{b}(t_{j+1}) \ \tf{w}{b}^{-1} (t_{j})\right).
	\end{equation*}
	The regulated translational velocity along the ISA was computed from the spatial screw twist components using the formulas in~\cite{verduyn2023enhancingmotiontrajectorysegmentation}.
	
	After computing the traversed geometric progress $s(t_k)$ at each timesample $t_k$ using \eqref{eq:num_progress}, a linear interpolating function was defined that maps each temporal sample $\tf{w}{b}(t_k)$ to a corresponding geometric sample $\tf{w}{b}(s(t_k))$ using \textit{screw linear interpolation} (ScLERP)\cite{kavan2008geometric}, a generalization of SLERP to $SE(3)$. This interpolation function was then resampled to obtain the geometric trajectory $\tf{w}{b}(s_i)$ at an equidistant geometric progress interval $s_i = i \Delta s$. The size of the interval $\Delta s$ was implicitly determined by setting the total number of geometric samples per trajectory equal to 50. 
	%Consequently, each trajectory obtained a numerically different value for its geometric progress interval $\Delta s$.

	\subsubsection{Invariant descriptors: Mitigating Frame Dependency}
	To deal with variations in the world and body frame, we propose to classify the resulting geometric trajectories based on their similarity to reference geometric trajectories in a manner that is invariant to world and body frame variations. 
	
	% Several invariant similarity measures exist in the literature, including the measures based on the DHB~\cite{Lee2018}, eFS~\cite{Vochten2015}, ISA~\cite{DeSchutterJoris2010}, RRV~\cite{RRV2018} and DSRF~\cite{DSRF2018} trajectory descriptors. 
	
	%Table~\ref{tab:measure_comparison} summarizes these key high-level properties of the DHB, eFS, ISA, RRV, and DSRF approaches, and compares them to the properties of the proposed BILTS measure.
	%
	%\begin{table}[h!]
	%	\centering
	%	\renewcommand{\arraystretch}{1.3}
	%	\caption{Comparison of Invariant Similarity Measures}
	%	\begin{tabular}{lccc}
		%		\hline
		%		\textbf{Approach} & \textbf{Left-invariance} & \textbf{Right-invariance} & \textbf{Locality} \\ \hline
		%		RRV~\cite{RRV2018}                & \cmark                   & \xmark                    & \xmark \\
		%		DSRF~\cite{DSRF2018}                & \cmark                   & \xmark                    & \xmark \\
		%		DHB~\cite{Lee2018}                 & \cmark                   & \xmark                    & \cmark \\
		%		eFS~\cite{Vochten2015}                 & \cmark                   & \xmark                    & \cmark \\
		%		ISA~\cite{DeSchutterJoris2010}                 & \cmark                   & \cmark                    & \cmark \\
		%		\textbf{BILTS}               & \cmark                   & \cmark                    & \cmark \\ \hline
		%	\end{tabular}
	%	\label{tab:measure_comparison}
	%\end{table}

	\subsection{Invariant Similarity: An Experimental Comparison}
	\label{sec:experimental_comparison}
	Each existing invariant similarity measure has its specific limitations. Common issues include sensitivity to noise near singularities, dependency on the choice of body reference point, and the requirement for prior motion segmentation. 
	% To address these limitations, we developed the novel BILTS measure. 
	
	We experimentally compared the performance of the BILTS measure to these approaches for the recognition of the object-manipulation tasks within the DLA and SYN datasets. For this comparison, a baseline recognition algorithm was used for all approaches, consisting of a \mbox{\textit{1-Nearest Neighbor}} (\mbox{1-NN}) classifier \cite{cover1967nearest} with DTW~\cite{sakoe1978dynamic}. 
	
	\subsubsection{Implementation of BILTS measure}
	We propose to define the similarity between two trajectories $\tf{w}{b}_1(s_k)$ and $\tf{w}{b}_2(s_k)$ as the average BILTS measure $\bar{d}_k$ between their corresponding trajectory shapes ${_f}\boldsymbol{Y}_1(s_k, m\Delta s)$ and  ${_f}\boldsymbol{Y}_2(s_k,m\Delta s)$, for $k$ ranging from $1$ to the total number of trajectory samples $N$: 
	\begin{align}
		\bar{d}_k ~\defeq~ \frac{1}{N}\sum_{k=1}^N
		\left\| {_f} \boldsymbol{Y}_1(s_k, m\Delta s) - {_f}\boldsymbol{Y}_2(s_k,m\Delta s) \right\|_W.
		\label{eq:globalmetric_discrete}
	\end{align}
	
	We computed both the unregularized (BILTS) and regularized (BILTS$^+$) versions of the discretized BILTS measure $d_k$. 
	
	When computing the BILTS measures, the geometric trajectories $\tf{w}{b}(s_k)$ were smoothed to filter out noise and minor human variations in the motions. The trajectory of the body’s reference point $\boldsymbol{p}^b(s_k)$ was smoothed using a Gaussian kernel, and the orientation trajectory, represented by quaternion coordinates $\boldsymbol{q}(s_k)$, was similarly smoothed. After smoothing, the quaternions were re-normalized to ensure unit length. The smoothed position and quaternion coordinates were then converted back to $4\times4$ homogeneous transformation matrices.
	
	From the smoothed geometric trajectories, we computed the spatial screw twists ${_w}\tw(s_k)$ at each trajectory sample using \eqref{eq:logm}. From these twists we computed the descriptor ${_w}\boldsymbol{Y}(s_k, m \Delta s)$ with \mbox{$m = \text{max}\left(1,\text{round}(\xi / \Delta s)\right)$} and with a tuned value for the progress scale $\xi$. We then computed the bi-invariant descriptor ${_f}\boldsymbol{Y}(s_k, m \Delta s)$ via the eQR-decomposition of ${_w}\boldsymbol{Y}(s_k, m \Delta s)\boldsymbol{A}$, as detailed in Section~\ref{sec:discrete}.
	
	In practice, the trajectory shapes in \eqref{eq:globalmetric_discrete} are not perfectly aligned in progress $s_k$. To address this, a progress alignment algorithm like DTW should be applied before computing~\eqref{eq:globalmetric_discrete}. For applications with stringent computational speed requirements, it is essential that the distance metric within DTW be efficient to evaluate. To address this, we propose to calculate the $3 \times 3$ diagonal matrices $\boldsymbol{\Sigma}_1$ and $\boldsymbol{\Sigma}_2$, which consist of the singular values of the upper and lower $3 \times 3$ matrices in ${_f}\boldsymbol{Y}(s_k, m\Delta s)$. The resulting six singular values provide a reduced but informative representation of the trajectory shape ${_f}\boldsymbol{Y}(s_k, m\Delta s)$, compressing it from 18 elements to 6. The trajectory shapes in \eqref{eq:globalmetric_discrete} can then be aligned along the progress $s_k$ by locally matching their singular value representations. After this DTW alignment, the average BILTS measure $\bar{d}_k$ between the aligned trajectory shapes was evaluated using~\eqref{eq:globalmetric_discrete}.
	
	\subsubsection{Implementation of other local invariant measures}
	We evaluated the performance of the BILTS measure to other similarity measures based on existing invariant trajectory descriptors. In this section, we focus on the similarity measures, $d^{DHB}$, $d^{\hspace{1pt}eFS}$, and $d^{ISA}$, based on the respective descriptors DHB, eFS and ISA. To ensure a fair comparison with the BILTS measure, we also introduced two tunable parameters, $L$ and $\lambda$, for these measures. These measures consist of the weighted euclidean norm of the elementwise differences between descriptors:
	\begin{equation*}
		\small 
			d^{DHB} = \left \Vert \begin{matrix} 
				L \Delta \omega \\ 
				L \lambda \Delta \theta_{\omega,1} \\
				L \lambda \Delta \theta_{\omega,2} \\
				\Delta v \\
				L \lambda \Delta \theta_{v,1} \\
				L \lambda \Delta \theta_{v,2} \end{matrix} \right \Vert , 
			d^{\hspace{1pt}eFS} = \left \Vert \begin{matrix} 
				L \Delta \omega \\ 
				L \lambda \Delta \omega_{\omega,1} \\
				L \lambda \Delta \omega_{\omega,2} \\
				\Delta v \\
				L \lambda \Delta \omega_{v,1} \\
				L \lambda \Delta \omega_{v,2} \end{matrix} \right \Vert , 
			d^{ISA} = \left \Vert \begin{matrix} 
				L \Delta \omega_1 \\ 
				L \lambda \Delta \omega_{2} \\
				L \lambda \Delta \omega_{3} \\
				\Delta v_1 \\
				\lambda \Delta v_2 \\
				\lambda \Delta v_3 \end{matrix} \right \Vert,
	\end{equation*}
	where $L$ is a parameter expressed in meters to weight the rotational components with the translation components. The hyperparameter $\lambda$ is introduced to weight the descriptor elements describing the first-order kinematics of the moving body with the ones describing the first-order kinematics of the functional frame(s). The kinematics of the functional frame(s) are known to be highly sensitive to noise near singularities and, for the eFS and ISA descriptors, can reach unbounded magnitudes. Hence, the introduction of this hyperparameter
	$\lambda$ is beneficial for these approaches, since it allows to directly control
	the influence of these kinematics during similarity measurement. Remark that the proposed BILTS measure provides a more physics-inspired approach for this weighting, related to a chosen progress scale for the trajectory shapes ($\delta s$ for the continuous-time case and $\xi$ for the discretized case). 
	This was explicitly shown in \eqref{eq:def_Y}, where the columns of $\boldsymbol{R}(s)$ are weighted with the Taylor series coefficients in $\boldsymbol{C}(\delta s)$ to obtain the continuous-time bi-invariant descriptor ${_f}\boldsymbol{\hat{Y}}(s,\delta s)$. 
	
	The DHB, eFS, and ISA descriptors and corresponding similarity measures require the estimation of higher-order trajectory derivatives. Hence, we implemented a linear Kalman smoother with a white-noise jerk-derivative model to simultaneously smooth the geometric trajectories, represented by position and quaternion coordinates, and estimate their corresponding higher-order derivatives. The quaternion coordinates were again re-normalized after smoothing to ensure unit length. The rotational velocity vector $\boldsymbol{\omega}(s_k)$ and its derivatives were subsequently calculated from the quaternion coordinates and their derivatives by applying the relations defined in~\cite{graf2008quaternionsdynamics}. Afterwards, the eFS and DHB descriptors were calculated using the analytical formulas in \cite{Vochten2015,Lee2018}. To calculate the ISA descriptor, the spatial screw twist $_w\tw(s_k)$ and its derivatives were first computed from the resulting rotational velocity vector,  the smoothed position coordinates, and their estimated derivatives, through the use of adjoint transformations that map body-frame quantities to the spatial frame \cite{lynch2017modern}. Afterwards, the analytical formulas in \cite{DeSchutterJoris2010} were implemented.
	
	Additionally, we alternatively computed the ISA descriptor by solving an Optimal Control Problem\cite{vochten2023invariant}, as this increases the descriptor's robustness to noise and singularities. We refer to this approach as the \textit{ISA-ocp} approach. To assess the improvement of the OCP approach compared to the analytical approach, the same smoothed geometric trajectories were used as input. The trajectory reconstruction tolerances within the OCPs were set to 2mm and 2$\degree$ according to \cite{vochten2023invariant}.
	
	For each approach (eFS, DHB, ISA, ISA-ocp), the similarity between two complete trajectories was then quantified as the average local shape similarity after DTW-alignment of the trajectories, using the above similarity measures as the distance metrics within the DTW algorithm. 
	
	\subsubsection{Implementation of other non-local invariant measures}
	We also evaluated the performance of the BILTS measure against the non-local similarity measures: RRV and DSRF. For these measures, we also first transformed the temporal rigid-body trajectories to geometric trajectories. Afterwards, the geometric trajectories were smoothed using Gaussian kernels, similarly to the BILTS approach.
	Then, the DSRF and RRV descriptors and measures, together with their corresponding DTW alignment approaches, were implemented as explained in \cite{DSRF2018,RRV2018}. These works do not explicitly introduce weighting parameters (such as $L$ or $\lambda$) within their motion similarity measures. Both RRV and DSRF approaches rescale the motion segment such that the total traversed arclength of the path traced by the body reference point is equal to one.  After this rescaling, the rotational components within the descriptors are implicitly weighted with respect to the translational components using a unit weighting factor. Since the RRV and DSRF approaches do not describe first-order kinematics of a local functional frame, we did not introduce the hyperparameter $\lambda$ for these approaches in our recognition experiments.

	\subsection{Parameter tuning}
	To compare the recognition performances of the different approaches, a structured \textit{training} and \textit{testing} routine was followed. For each motion class, the invariant descriptors of the trials from one context (referred to as the \textit{reference context}) were computed and retained. These descriptors were designated as the reference descriptors for the respective motion class. We utilized reference descriptors from only one context to assess the effectiveness of the different approaches when extrapolating to unrepresented contexts. The contexts \textit{Original} and \textit{Normal: seen from viewpoint 2} served as the reference contexts within the SYN and DLA datasets, respectively. 
	
	The computation of the invariant similarity measures based on their corresponding invariant descriptors required the tuning of the following parameters: 
	\begin{itemize}
		\item $L$ and $\lambda$ for the DHB, eFS, ISA, and ISA-ocp measures,
		\item $L$ and $\xi$ for the BILTS and BILTS$^+$ measures,
		\item $L$ for the RRV and DSRF measures solely when using the screw path as the geometric progress, since this progress type requires a factor $L$ to weight the rotational velocity against the translational velocity along the ISA.
	\end{itemize} 
	Optimal values for these parameters were identified through training on subsets of the SYN and DLA datasets, referred to as their respective \textit{training sets}. These training sets consisted of the first two trials of each context, excluding the reference context. These trials were then classified based on their similarity to the trials from the reference context using a 1-NN classifier. Optimal values for the parameters were identified through a grid search over a predefined range. Specifically, the parameter values resulting in the highest recognition rate across this grid were considered the optimal parameters for the training set and were retained. Hence, the highest recognition rate served as an objective criterion for parameter training. After training, the remaining trials within the datasets (\textit{test trials}) were classified using the same 1-NN classifier.  
	
	\subsection{Experimental results}
	
	The parameter values and corresponding recognition rates (RR) are reported in Tables~\ref{tab:validation_results1}, \ref{tab:validation_results2}, and \ref{tab:validation_results3}. The first column lists the invariant similarity measurement approaches. The second column specifies the type of geometric progress. Tables~\ref{tab:validation_results1} and \ref{tab:validation_results2} show the highest RR and corresponding parameter values obtained during training. Table~\ref{tab:validation_results3} shows the achieved RR during testing, with the highest RR in each column depicted in bold. The fifth and sixth columns show the difference in RR relative to the one for the original DLA dataset. The final column shows the average RR and standard deviation across all four datasets. The average results for the proposed BILTS$^+$ measure is highlighted in green.
	
		% Define custom colors
	\definecolor{mygreen}{RGB}{34, 139, 34}
	
	\sethlcolor{mygreen}
	\definecolor{lightgray}{RGB}{235, 235, 255}
	\definecolor{darkgray}{RGB}{170, 170, 255}
	\definecolor{darkgreen}{RGB}{50, 220, 50}

\subsection{Interpretation of the results}
Table~\ref{tab:validation_results3} shows that the regularized BILTS$^+$ measure achieved the \mbox{\textbf{highest average RR}} of 95.3\%, outperforming on average all other similarity measures across every geometric progress type. Its robustness is further underscored by its \mbox{\textbf{lowest sensitivity to contextual variations}}, reflected by its lowest standard deviation of 3.4\%. 
%Specifically, BILTS$^+$ obtained an average RR of $95.3~(\pm3.4)$\%. 

\begin{table}[t]
	\centering
	\caption{Training results (\%) for the Synthetic and original DLA datasets.}
	\label{tab:validation_results1}	
	\resizebox{\linewidth}{!}{%
		\renewcommand{\arraystretch}{0.9}
		\begin{tabular}{llcc>{\columncolor{lightgray}}cccc>{\columncolor{lightgray}}c}
			\toprule 
			\textbf{Approach} & \textbf{Progress} &
			\multicolumn{3}{c}{\textbf{SYN dataset}} &&  \multicolumn{3}{c}{\textbf{DLA dataset}} \\
			\addlinespace
			&& \multicolumn{3}{c}{\textbf{synthetic elementary}} && \multicolumn{3}{c}{\textbf{recordings of daily-life}}\\
			&& \multicolumn{3}{c}{\textbf{rigid-body motions}} && \multicolumn{3}{c}{\textbf{object-manipulation tasks}} \\
			&&&&\cellcolor{white}  &&&&\cellcolor{white}\\ 
			\cmidrule(lr){3-5} \cmidrule(lr){7-9}
			\addlinespace
			&& \multicolumn{2}{c}{\textbf{param. values}} & \textbf{RR (\%)} &&
			\multicolumn{2}{c}{\textbf{param. values}} &  \textbf{RR (\%)} \\
			&&\multicolumn{1}{c}{$L$}&\multicolumn{1}{c}{$\lambda$}&\textbf{training}&&\multicolumn{1}{c}{$L$}&\multicolumn{1}{c}{$\lambda$}&\textbf{training}\\
			\addlinespace
			& arclength & $0.3$m & $100$&  75.0 & & $0.1$m & $10$ &  87.6 \\
			DHB~\cite{Lee2018} & angle & $ 0.3$m & $ 100$ &  55.4 & & $ 0.9$m & $  0.1$ &  93.5\\
			& screw path & $ 0.3$m & $  100$ &  65.7 & & $ 0.5$m & $  0.01$ &  95.5 \\
			\addlinespace
			& arclength & $ 0.1$m & $  10$ &  61.4 & & $0.1$m & $0.1$ &  82.1  \\
			eFS~\cite{Vochten2015} & angle & $ 0.7$m & $0.1$ &  52.9  & & $ 0.7$m & $  10^{-3}$ &  90.0 \\
			& screw path & $ 0.1$m & $  10$ &  65.7 & & $ 0.5$m & $  0.01$ &  96.0 \\
			\addlinespace
			\multirow{2}{*}{ISA~\cite{DeSchutterJoris2010}} & angle & $ 0.1$m & $  10^{-3}$ &  62.9 & & $ 0.5$m & $  10^{-4}$ & 90.5 \\
			& screw path & $ 0.3$m & $  0.01$ &  66.1 & & $ 0.5$m & $  10^{-4}$ &  92.0 \\
			\addlinespace
			ISA-ocp~\cite{vochten2023invariant} & screw path & $ 0.3$m & $  1$ &  85.7 & & $0.3$m & $10^{-4}$ &  93.0 \\
			\addlinespace
			\cmidrule(lr){3-5} \cmidrule(lr){7-9}
			&& \multicolumn{1}{c}{$L$} & \multicolumn{1}{c}{$\xi$} &\cellcolor{white}&& \multicolumn{1}{c}{$L$} & \multicolumn{1}{c}{$\xi$}&\cellcolor{white}\\
			\addlinespace
			\multirow{2}{*}{BILTS} & angle & $ 0.9$m & $  10\degree$ &  72.5 & & $ 0.3$m & $  20\degree$ & 93.5 \\
			& screw path & $ 0.3$m & $  0.06$m &  75.0 & & $ 0.7$m & $  0.15$m &  95.0 \\
			\addlinespace
			\textbf{BILTS$^+$} & screw path & $ 0.7$m & $  0.12$m &  100& & $ 0.3$m & $0.03$m &  95.5\\ 
			\addlinespace
			\cmidrule(lr){3-5} \cmidrule(lr){7-9}
			&& \multicolumn{1}{c}{$L$} &&\cellcolor{white}&&\multicolumn{1}{c}{$L$}&\cellcolor{white}\\ 
			\addlinespace
			RRV~\cite{RRV2018} & screw path & $ 0.9$m &  &  53.6 & & $ 0.9$m &  &  92.5 \\
			\addlinespace
			DSRF~\cite{DSRF2018} & screw path & $ 0.9$m &  &  88.2 & & $ 0.5$m &  & 99.0  \\
			\addlinespace
			\bottomrule
		\end{tabular}
	}
	\vspace{-3pt}
\end{table}
% \vspace{-22pt}
\begin{table}[t]
	\centering
	\caption{Training results (\%) for the adapted DLA datasets.}
	\label{tab:validation_results2}	
	\resizebox{\linewidth}{!}{%
		\renewcommand{\arraystretch}{0.9}
		\begin{tabular}{llcc>{\columncolor{lightgray}}cccc>{\columncolor{lightgray}}c}
			\toprule 
			\textbf{Approach} & \textbf{Progress} &
			\multicolumn{3}{c}{\textbf{adapted DLA 1}} &&  \multicolumn{3}{c}{\textbf{adapted DLA 2}} \\
			\addlinespace
			&& \multicolumn{3}{c}{\textbf{artificial change in}} && \multicolumn{3}{c}{\textbf{successive motions}}\\
			&& \multicolumn{3}{c}{\textbf{body reference point}} && \multicolumn{3}{c}{\textbf{in varied directions}} \\
			&&&&\cellcolor{white}  &&&&\cellcolor{white}\\ 
			\cmidrule(lr){3-5} \cmidrule(lr){7-9}
			\addlinespace
			&& \multicolumn{2}{c}{\textbf{param. values}} & \textbf{RR (\%)} &&
			\multicolumn{2}{c}{\textbf{param. values}} &  \textbf{RR (\%)} \\
			&&\multicolumn{1}{c}{$L$}&\multicolumn{1}{c}{$\lambda$}&\textbf{training}&&\multicolumn{1}{c}{$L$}&\multicolumn{1}{c}{$\lambda$}&\textbf{training}\\
			\addlinespace
			& arclength & $0.1$m & $10$&  65.2 & & $0.1$m & $10$ &  78.1 \\
			DHB~\cite{Lee2018} & angle & $ 0.9$m & $ 0.1$ &  76.6 & & $ 0.7$m & $  0.1$ &  92.0\\
			& screw path & $ 0.5$m & $  1$ &  80.6 & & $ 0.5$m & $  0.01$ &  96.5 \\
			\addlinespace
			& arclength & $ 0.1$m & $  0.1$ &  52.7 & & $ 0.1$m & $  10^{-3}$ &  74.6  \\
			eFS~\cite{Vochten2015} & angle & $ 0.9$m & $  0.01$ &  71.6  & & $ 0.9$m & $  10^{-3}$ &  91.0 \\
			& screw path & $ 0.7$m & $0.01$ &  77.1 & & $ 0.5$m & $  10^{-3}$ &  96.5 \\
			\addlinespace
			\multirow{2}{*}{ISA~\cite{DeSchutterJoris2010}} & angle & $ 0.5$m & $  10^{-4}$ &  91.0 & & $ 0.3$m & $  10^{-3}$ & 93.0 \\
			& screw path & $ 0.5$m & $  10^{-4}$ &  92.5 & & $ 0.7$m & $  10^{-4}$ &  94.0 \\
			\addlinespace
			ISA-ocp~\cite{vochten2023invariant} & screw path & $ 0.3$m & $0.01$ &  92.0 & & $0.3$m & $10^{-4}$ &  79.6\\
			\addlinespace
			\cmidrule(lr){3-5} \cmidrule(lr){7-9}
			&& \multicolumn{1}{c}{$L$} & \multicolumn{1}{c}{$\xi$} &\cellcolor{white}&& \multicolumn{1}{c}{$L$} & \multicolumn{1}{c}{$\xi$}&\cellcolor{white}\\
			\addlinespace
			\multirow{2}{*}{BILTS} & angle & $ 0.3$m & $  20\degree$ &  93.0 & & $ 0.9$m & $  30\degree$ & 91.0 \\
			& screw path & $ 0.7$m & $  0.15$m &  95.0 & & $ 0.9$m & $  0.03$m &  88.1 \\
			\addlinespace
			\textbf{BILTS$^+$} & screw path & $ 0.7$m & $  0.15$m &  94.0& & $ 0.1$m & $0.03$m &  93.0\\ 
			\addlinespace
			\cmidrule(lr){3-5} \cmidrule(lr){7-9}
			&& \multicolumn{1}{c}{$L$} &&\cellcolor{white}&&\multicolumn{1}{c}{$L$}&\cellcolor{white}\\ 
			\addlinespace
			RRV~\cite{RRV2018} & screw path & $ 0.9$m &  &  85.1 & & $ 0.9$m &  &  65.7 \\
			\addlinespace
			DSRF~\cite{DSRF2018} & screw path & $ 0.7$m &  &  96.0 & & $ 0.1$m &  & 73.6  \\
			\addlinespace
			\bottomrule
		\end{tabular}
	}
	\vspace{-5pt}
\end{table}

\subsubsection{Interpretation of trained parameter values}

Tables~\ref{tab:validation_results1} and \ref{tab:validation_results2} show that the hyperparameter $\lambda$ obtained relatively high values for the DHB approach compared to the eFS, ISA and ISA-ocp approaches. The reason for this is that the corresponding DHB invariants do not represent the instantaneous kinematics of the functional FS frames, but instead encode the discrete angles between successive FS frames along the trajectory. Hence, on a dense grid --here, consisting of 50 trajectory samples per trajectory-- the DHB angles became relatively small. Therefore, to maintain a significant influence within the similarity measure, $\lambda$ had to be set to a higher value to amplify the impact of these angles. 

For the ISA approach, the hyperparameter $\lambda$ obtained significantly low values. Consequently, substantially downscaling the moving frame invariants --representing the kinematics of the functional frame within the ISA measure-- improved recognition accuracy during training. This downscaling makes sense, since these invariants are highly sensitive to noise and can obtain large unbounded values. In contrast, the \mbox{ISA-ocp} approach obtained larger values for $\lambda$ for the SYN dataset, likely due to the improved robustness of these invariants when computed by solving an OCP~\cite{Vochten2015,vochten2023invariant}. 
	
%	In case of the BILTS approach, the progress scale $\xi$ obtained the highest values during training for the regularized case on the SYN dataset, with $\xi = 40\degree$ when using \textit{angle}, and \mbox{$\xi = 15$cm} when using \textit{screw path} as the geometric progress. This indicates that a larger progress scale was advantageous for accurately recognizing the elementary motions within the SYN dataset. This is expected, since distinguishing between circular and helical translations, or between fixed-axis and precession rotations, requires capturing higher-order shape properties that become more discernible over larger progress scales.
	
\subsubsection{Detailed results for the SYN dataset}
The DHB and eFS approaches demonstrate limited recognition accuracy on the SYN dataset, likely due to their dependency on the body reference point (lack of right-invariance), as the SYN dataset contains substantial variations in the body reference frames. Additionally, both DHB and eFS were unable to distinguish screw motions with positive pitch from ones with negative pitch due to the loss of internal kinematics between rotation and translation. Furthermore, for pure translations (linear, circular, and helical), the FS frames for orientation were ill-defined and hence highly sensitive to noise, which further impacted recognition accuracy.

The unregularized ISA approach also demonstrated limited recognition accuracy, likely due to the impact of singularities. Six of the seven elementary rigid-body motions in the SYN dataset present cases where the ISA invariants are ill-defined. Specifically, for pure translations, the entire functional frame is ill-defined. Furthermore, for motions with a fixed rotation axis, the orientation of the $y$- and $z$-axes of the moving frame and the position of the moving frame along this axis are ill-defined. The unregularized BILTS approach also demonstrated limited recognition accuracy due to these singularities. However, the BILTS descriptor elements remained bounded even in these singular cases, unlike those in the ISA descriptor, leading to relatively higher recognition accuracies.

The RRV approach also showed limited recognition accuracy. Like the DHB and eFS approaches, the RRV descriptor lacks right-invariance. Additionally, it encounters similar singularities as the ISA and BILTS approaches, as the functional frame of the RRV descriptor is determined by the left singular vectors of a \mbox{$3\times N$} matrix consisting of a sequence of $N$ rotation axis vectors corresponding to the body's orientation trajectory. These left singular vectors are ill-defined for pure translations. Also, only the first singular vector remains well-defined in the case of fixed-axis rotations and pure screw motions. 
The lack of right-invariance and this ambiguity in singular cases highly impacted RRV's recognition accuracy.

The DSRF approach achieved relatively high recognition rates on the SYN dataset, benefiting from a robust spatial frame orientation alignment method. However, it did not achieve a 100\% recognition rate due to its lack of right-invariance. Additionally, for pure rotations about a fixed rotation axis where the reference point lies on this axis, normalizing the reference point trajectory to unit length introduced ambiguity and severe noise sensitivity in the rescaled position trajectory, thereby hindering accurate trajectory recognition.

As expected, the regularized approaches ISA-ocp and BILTS$^+$ performed significantly better than their unregularized counterparts for the SYN dataset.  The ISA-ocp approach demonstrated improved recognition accuracy. However, the OCP's solver occasionally failed to converge consistently to repeatable solutions, which affected the overall recognition accuracy. The regularized BILTS$^+$ measure showed exceptional robustness to singularities, achieving an impressive RR of 100\%. This confirms that the proposed regularization actions within BILTS$^+$ effectively handled trajectory-shape singularities, ensuring they no longer compromise accurate trajectory recognition.

\begin{table}[t]
	\centering
	\caption{Test results (\%) for SYN, DLA and adapted DLA datasets.}
	\label{tab:validation_results3}	
	\resizebox{\linewidth}{!}{%
		\renewcommand{\arraystretch}{0.9}
		\begin{tabular}{llllll>{\columncolor{lightgray}}l}
			\toprule 
			\textbf{Approach} & \textbf{Progress} &
			\textbf{SYN} & \textbf{DLA} &  \textbf{adapted 1} & \textbf{adapted 2} &  \textbf{mean ($\pm \sigma$)}\\
			\addlinespace
			& arclength & 72.6 & 83.9 & 61.5~\textcolor{red}{(-22.4)} & 77.3~\textcolor{red}{(-6.6)} & 73.8 $(\pm9.4)$\\
			DHB~\cite{Lee2018} & angle & 58.8 & 91.4 & 75.2~\textcolor{red}{(-16.2)} & 89.4~\textcolor{red}{(-2.0)} & 78.7 $(\pm15.1)$\\
			& screw path & 68.6 & 92.4 & 78.1~\textcolor{red}{(-14.3)} & 94.0~\textcolor{mygreen}{(1.6)} & 83.3 $(\pm12.1)$\\
			\addlinespace
			& arclength &  62.3 & 77.8 & 51.6~\textcolor{red}{(-26.2)} & 73.9~\textcolor{red}{(-3.9)} & 66.4 $(\pm11.9)$\\
			eFS~\cite{Vochten2015} & angle & 57.5 & 90.7 & 70.9~\textcolor{red}{(-19.8)} & 89.2~\textcolor{red}{(-1.5)} & 77.1 $(\pm15.9)$\\
			& screw path & 66.8 & 93.5 & 75.5~\textcolor{red}{(-18.0)} & 93.9~\textcolor{mygreen}{(0.4)} & 82.4 $(\pm13.5)$\\
			\addlinespace
			\multirow{2}{*}{ISA~\cite{DeSchutterJoris2010}} & angle & 61.6 & 90.3 & 90.3~(-0.0) & 91.0~\textcolor{mygreen}{(0.7)} & 83.3 $(\pm14.5)$\\
			& screw path & 68.3 & 91.0 & 89.9~\textcolor{red}{(-1.1)} & 91.5~\textcolor{mygreen}{(0.5)} & 85.2 $(\pm11.3)$\\
			\addlinespace
			ISA-ocp~\cite{vochten2023invariant} & screw path & 85.5 & 91.9 & 90.4~\textcolor{red}{(-1.5)} & 77.8~\textcolor{red}{(-14.1)}  & 86.4 $(\pm6.4)$\\
			\addlinespace
			\multirow{2}{*}{BILTS} & angle & 70.0 & 91.3 & 91.4~\textcolor{mygreen}{(0.1)} & 83.1~\textcolor{red}{(-8.2)}  & 84.0 $(\pm10.1)$\\
			& screw path & 71.0 & 91.8 & 92.1~\textcolor{mygreen}{(0.3)} & 82.8~\textcolor{red}{(-9.0)}  & 84.4 $(\pm9.9)$\\
			\addlinespace
			\textbf{BILTS$^+$} & screw path & \textbf{100} & 92.6 & \textbf{93.2}~\textcolor{mygreen}{(0.6)} & \textbf{95.1}~\textcolor{mygreen}{(2.5)}  & \cellcolor{darkgreen} $\boldsymbol{95.3~(\pm3.4)}$\\
			\addlinespace
			& arclength & 59.5 & 94.4 & 77.8~\textcolor{red}{(-16.6)} & 56.0~\textcolor{red}{(-38.4)}  & 71.9 $(\pm17.8)$\\
			RRV~\cite{RRV2018} & angle &  59.8 & 90.5 & 80.7~\textcolor{red}{(-9.8)} & 60.7~\textcolor{red}{(-29.8)}  & 72.9 $(\pm15.2)$\\
			& screw path & 54.6 & 90.4 & 80.2~\textcolor{red}{(-10.2)} & 67.0~\textcolor{red}{(-23.4)}  & 73.1 $(\pm15.6)$\\
			\addlinespace
			& arclength & 89.4 & \textbf{98.8} & 91.3~\textcolor{red}{(-7.5)} & 72.7~\textcolor{red}{(-26.1)}  & 88.1 $(\pm11.0)$\\
			DSRF~\cite{DSRF2018} & angle & 89.8 & 95.3 & 91.0~\textcolor{red}{(-4.3)} & 74.1~\textcolor{red}{(-21.2)}  & 87.6 $(\pm9.3)$\\
			& screw path & 88.2 & 96.2 & 92.3~\textcolor{red}{(-3.9)} & 72.1~\textcolor{red}{(-24.1)}  & 87.2 $(\pm10.6)$\\
			\addlinespace
			\bottomrule
		\end{tabular}
	}
	\vspace{-10pt}
\end{table}

\subsubsection{Detailed results for the original DLA dataset}

The RRV and DSRF measures achieved notably high recognition ratios (94.4\% for RRV and an impressive 98.8\% for DSRF) for the DLA dataset. These measures, which lack right-invariance, are dependent on the choice of body reference point—a dependency that proved non-detrimental for this specific dataset. Likely, the body reference point (calculated as the average of visible LEDs) remained relatively consistent across trials, explaining that this dependency did not hinder accurate motion recognition. Additionally, the RRV and DSRF descriptors, which lack true locality, require that the motion be consistently segmented, which was the case for this dataset. This consistency in segmentation likely contributed to DSRF’s high performance, also suggesting that a focus on true locality offers limited benefit for the original DLA dataset.

The regularized BILTS$^+$ and ISA-ocp measures improved compared to their unregularized counterparts, but achieved only the seventh and eighth highest recognition ratios (92.6\% and 91.9\%, respectively) for the original DLA dataset. Further analysis revealed that these results were impacted by confusion between the \textit{scooping followed by pouring} and \textit{scooping} motion classes, which both locally involve a \textit{scooping} motion. 

Above confusion reflects the challenge of recognizing motions purely based on local bi-invariant trajectory shapes without contextual cues such as the history of the motion, and the location and orientation of the body and world reference frames. 
%This challenge becomes particularly more pronounced when differentiating between subtly varied motion classes.  
Accordingly, we restate that the problem this paper addresses is dealing with scenarios in which these contextual cues significantly differ among motions that should be categorized under a single motion class. In this case, and only in this case, the removal of these contextual dependencies enhances accuracy in motion recognition. To demonstrate this enhanced accuracy more rigorously, we developed the adapted DLA datasets which include extra contextual variations.

% Additionally, this classification confusion was most notable in trials from the first context (\textit{normal: viewpoint 1}), where invariant descriptors showed greater variability than in later contexts. This suggests that only the later-recorded \textit{scooping} motions were more consistently executed, likely due to the demonstrator's habituation over time.

	\subsubsection{Advantage of right-invariance}
	The adapted DLA 1 dataset was designed to underscore the drawbacks of approaches that lack right-invariance. Table~\ref{tab:validation_results3} shows that the implemented adaptation of the body reference point affected the recognition accuracy of all approaches, including those that are right-invariant, due to the trajectory smoothing as a preprocessing step.  
	Specifically, the reference point trajectories were smoothed prior to descriptor calculation. 
	Hence, changing the location of this reference point changed the kinematics of its trajectory, thereby introducing minor reference point dependencies after smoothing. 
	
	Nevertheless, the ISA, BILTS and BILTS$^+$ approaches demonstrated significantly lower sensitivity to these reference point variations compared to the ones that lack right-invariance (DHB, eFS, RRV, DSRF), with variations in RR less than 2\%. 
	%The ISA-ocp approach exhibited a slightly higher sensitivity, although still limited. This increased sensitivity can be attributed to the ISA-ocp approach's minimization of the reconstruction error on the trajectory of the reference point, which introduces again a slight dependency on this reference point.
	
	\subsubsection{Advantage of locality and left-invariance}
	The adapted DLA 2 dataset was designed to underscore the drawbacks of approaches that lack left-invariance and true locality. Table~\ref{tab:validation_results3} shows that the added complexity --successively executing motions in diverse directions-- again impacted the recognition accuracy of all approaches, including those that are left-invariant and local. This effect arose because the concatenation of motions introduced random transition effects, which impacted the recognition accuracy across all approaches. 
	
	Nevertheless, the local approaches (DHB, eFS, ISA, BILTS and BILTS$^+$) demonstrated significantly lower decrement in accuracy compared to those lacking true locality (RRV, DSRF). Somewhat unexpectedly, the ISA-ocp approach, despite being based on the local ISA descriptor, showed a relatively high accuracy decrement. This can be attributed to the \mbox{ISA-ocp's} regularization strategy, which penalizes high magnitudes of the moving frame invariants. This likely `smeared out' the peaks in the moving frame invariants caused by the introduced transition effects, thereby amplifying their negative impact on the overall recognition accuracy.
	
	The BILTS$^+$ measure demonstrated a surprisingly high RR of 95.1\% on the adapted DLA dataset 2, which an increment of 2.5\% compared to the RR on the original DLA dataset. This performance could be explained by the higher robustness of the BILTS measure due to the proposed regularization actions, combined with the artificial repetition of each motion per trial in this adapted dataset. This repetition may have enhanced the classifier's ability to capture subtle variations between similar motions, resulting in an improved the recognition rate.
	
	\section{Discussion and conclusion}
	\label{sec:discussion_conclusion}
	
	\subsection{Research contributions}
	We introduced the novel continous-time and discretized \textit{Bi-Invariant Local Trajectory-Shape Similarity} (BILTS) measures for rigid-body motion, which quantify the difference in trajectory shape as a single scalar value ($d$ or $d_k$) in a bi-invariant manner. The definition of the continuous-time measure is based on the bi-invariant trajectory-shape descriptor ${_f}\boldsymbol{\hat{Y}}(s,\delta s)$, which describes the third-order shape of the trajectory in a bi-invariant manner. This was achieved by expressing second-order Taylor series approximation of the spatial screw twist ${_w}\tw(s)$ in the functional bi-invariant frame $\{f\}$, which is uniquely defined by the ISA and its first-order kinematics. 
	
	Hence, we separated out the context (dependencies on the spatial frame $\{w\}$ and body frame $\{b\}$) from the body's intrinsic motion. We achieved this without needing to consider the body's specific type, shape or inertial properties.
	%Hence, we removed the dependency on the spatial frame $\{w\}$ and body frame $\{b\}$ using solely information of the body's kinematics, without having to consider the specific type, shape or inertial properties of this body. 
	A practical advantage of this approach is that it does not require the careful selection and precise calibration of reference frames when comparing motions. Additionally, the BILTS measure is capable of generalizing motions across different object types, enabling the detection of similarity between motions performed with various objects.
	% (e.g., pouring motions performed with a \textit{cup}, \textit{jug}, or \textit{kettle}).
	
	Separating out the context is useful when the context is variable and is considered a disturbance that hinders effective motion recognition. When the following two conditions hold simultaneously, the BILTS measure is particularly well-suited:
	(1) There are significant variations in both the body and world reference frames across trials that are intended to be recognized as part of the same motion class. (2) The motion trajectories are not segmented or inaccurately segmented.
	
	Analytical relations were then derived between the continuous-time BILTS measure and other similarity measures based on existing invariant descriptors (FS, eFS and ISA). Compared to these measures, the BILTS measure is the only one that posesses both bi-invariance and boundedness. It is also the only one that captures full third-order shape identity.  
	
	We then introduced the discretized BILTS measure $d_k$ and explained how to compute it from discrete-time rigid-body trajectories, without requiring explicit estimation of higher-order trajectory derivatives. Lastly, we introduced regularization to increase the BILTS measure's robustness near singularities.
	
	We aimed to make the use of BILTS straightforward by keeping the set of parameters limited and intuitive. These parameters include the geometric scale factor $L$, and the progress scale $\xi$. The selection of $\xi$ involves balancing between a more localized (small $\xi$) and a more global (large $\xi$) trajectory-shape descriptor, where a larger $\xi$ is beneficial for trajectory recognition where smaller local shape deviations obscure the more global trajectory shape. The parameter $L$ governs the weighting between rotational and translational components of the moving body. In the case of regularization, it also defines the spherical region around the origin of $\{b\}$ within which the origin of $\{f\}$ is constrained. In the experiments, the values for the parameters $L$ and $\xi$ were determined through tuning on training sets, where the highest recognition rate during training served as an objective criterion for this tuning.
	
	Due its boundedness, the unregularized BILTS measure already provides superior robustness against singularities compared to the unregularized ISA measure. This was demonstrated in motion recognition experiments, where BILTS outperformed ISA in recognizing synthetic elementary rigid-body motions which were rich in singularities. 
	
	Moreover, the regularized BILTS$^+$ measure demonstrated exceptional robustness to singularities and contextual variations, achieving a 100\% recognition rate on the SYN dataset. On average, BILTS$^+$ also attained the highest recognition ratio of 95.3\% compared to other existing similarity measures (DHB, eFS, ISA, ISA-ocp, RRV, DSRF). It also exhibited the least sensitivity to contextual variations, shown by its lowest standard deviation of 3.4\% across the various datasets.

	%The mathematical relationships between the BILTS descriptor and the ISA descriptor are reported in Sec.~\ref{sec:relations}. Furthermore, relations with the Bishop and Frenet-Serret invariants are reported in App. \ref{app:fs_point} and \ref{app:bishop}.  
	
	%% bi-invariance is not that well-known in robotics community, so we might not have to make the next paragraph explicit
	% The dissimilarity measure based on the BILTS descriptor enables us to evaluate the similarity between trajectories in a bi-invariant manner. However, it is important to note that, while this measure is bi-invariant, it should not be interpreted as a bi-invariant distance metric on $SE(3)$. Such a metric is known not to exist \cite{loncaric1985geometrical,Park1995}. Instead, our measure focuses on comparing the shapes of rigid-body trajectories.
	
	\subsection{BILTS with reintroduced context for enhanced recognition}
	
	%We achieved the BILTS measure by representing the object's trajectory shape in a bi-invariant frame that is uniquely defined by the ISA and its first-order kinematics. Hence, we removed the dependency on the spatial frame $\{w\}$ and body frame $\{b\}$ without considering the specific type of object that is being used. A practical advantage of this approach is that it does not require the careful selection and precise calibration of reference frames when comparing motions. Additionally, the BILTS measure is capable of generalizing motions across different object types, enabling the detection of similarity between motions performed with various objects (e.g., pouring motions performed with a \textit{cup}, \textit{jug}, or \textit{kettle}).
	
	However, a limitation among invariant similarity measurement approaches is that when the context is not variable, removing contextual information will not improve recognition accuracy. Furthermore, invariant similarity measures may fail to distinguish motions that have a high invariant similarity but differ in their context. To effectively differentiate between such motions (which is outside the scope of this paper), preserving or reintroducing contextual information will be advantageous. 
	
	Another consideration relates to the achieved recognition accuracies within the experimental validation. We used the baseline \mbox{1-NN} classification algorithm for our experimental validation to reveal the root causes of limitations of existing similarity measures.  Nevertheless, BILTS can be combined with more advanced recognition algorithms from the literature to further enhance recognition accuracy. 
	
	Hence, future work is to explore the BILTS measure with reintroduced contextual information by integrating more advanced, learning-based algorithms. For example, a self-attention mechanism~\cite{selfattentionRecognition} could be added to additionally capture long-range contextual patterns in trajectory data. This approach would allow the detection of both local invariant shape features and long-range contextual features, without requiring a complex multi-layered network.

	\subsection{Other uses for the proposed BILTS measure}
	
	While Section~\ref{sec:recognition} focused on recognizing object-manipulation tasks, the proposed BILTS measure is also useful for other applications involving rigid-body trajectories. Several of these applications are outlined below.
	
	\vspace{3pt}
	\subsubsection{Application to real-time gesture recognition}
	
	The BILTS measure also serves as a valuable tool for gesture recognition in human-robot interaction applications. In particular, by representing the motion of the human hand as a rigid-body trajectory, the similarity between hand gestures can be measured in a bi-invariant manner using the proposed BILTS measure. This is  shown in a proof of concept involving real-time gesture recognition. A video showcasing this proof of concept is made accessible online\footnote{Available on Zenodo at \url{https://doi.org/10.5281/zenodo.14679070} .}.
	
	For this proof of concept, the user’s hand motion was captured by holding an HTC Vive Tracker. In real-time, the trajectory data from the tracker was continuously reparameterized using the screw-based geometric progress parameter~\cite{verduyn2023enhancingmotiontrajectorysegmentation}, and successive BILTS descriptor instances were computed on the fly. The following parameter values were used: a geometric progress resolution of $\Delta s = 0.01$m, a geometric scale of $L = 0.2$m, and a progress scale of $\xi = 0.1$m.
	
	The following seven gestures were defined a priori: \textit{Go Home, Go Up, Go Down, Go Left, Go Right, Open Gripper,} and \textit{Close Gripper}. These gestures involved rotational movements, translational movements, or a combination of both performed by the hand. Additionally, a stop signal --a fast rotation of the hand-- was implemented as an interruption signal. For each of the seven gestures, gesture templates were pre-recorded and the corresponding discretized BILTS descriptors were pre-computed.  
	
	Similarly to the recognition experiments described in Section~\ref{sec:experimental_comparison}, the robot recognized the hand gestures using a 1-NN classifier, with the regularized BILTS$^+$ measure serving as the distance measure.  Specifically, each pre-recorded gesture template was associated with a specific trajectory length, and based on this length, corresponding sliding windows of BILTS descriptor instances were defined. Each time a new BILTS descriptor instance was computed, the sliding windows were updated, and a candidate gesture was proposed by the 1-NN classifier. Candidate gestures were accepted when the BILTS$^+$ measure between the performed gesture and the closest neighbor within the 1-NN classifier remained below a threshold, effectively reducing false positives. Once accepted, the recognized gestures were mapped to motion plans in the robot's joint space, which were then executed by the robot.
	
	This proof of concept showcases the advantages of the BILTS measure for gesture recognition applications. The BILTS measure's invariance to both the world and body reference frames eliminates the need for calibration between the HTC Vive motion capture setup and the robot's base frame. This simplifies setup deployment and also allows users to move freely or hold the tracker in arbitrary ways without compromising recognition accuracy. 

	\vspace{3pt}
	\subsubsection{Application to trajectory segmentation}
	
	The BILTS measure could also serve as a valuable tool for trajectory segmentation in diverse contexts, since it allows to segment rigid-body trajectories by detecting sudden changes in the bi-invariant trajectory shape. Specifically, the change in shape between trajectory samples $\tf{w}{b}(s_k)$ and $\tf{w}{b}(s_{k+1})$ can be quantified as a scalar value $d_{k+1,k}$ using the BILTS measure between consecutive instances along the trajectory:
	\begin{equation}
		d_{k+1,k} = \left\| {_f} \boldsymbol{Y}(s_{k+1}, m\Delta s) - {_f}\boldsymbol{Y}_2(s_{k},m\Delta s) \right\|_W.
	\end{equation}
	
	Segments of homogeneous trajectory shape can then be found by searching for intervals where $d_{k+1,k}$ remains below a predefined threshold. Alternatively, discrete segmentation points can be found by locating the maxima of $d_{k+1,k}$. This method can be interpreted as an extension towards rigid-body trajectories of the geometric segmentation approach for point curves presented in \cite{SegmentationContour1988}. In \cite{SegmentationContour1988} breakpoints are detected by identifying the maxima of the total curvature along a curve.
	
	Due to the bi-invariance of $d_{k+1,k}$, the segmentation result will be bi-invariant. Consequently, changes in the body or world frames will not affect the segmentation outcomes, eliminating the need for accurate calibration of these frames to ensure consistent segmentation across diverse contexts.
	
\begin{figure}[t]
		\centering
		\includegraphics[width = 0.45\linewidth]{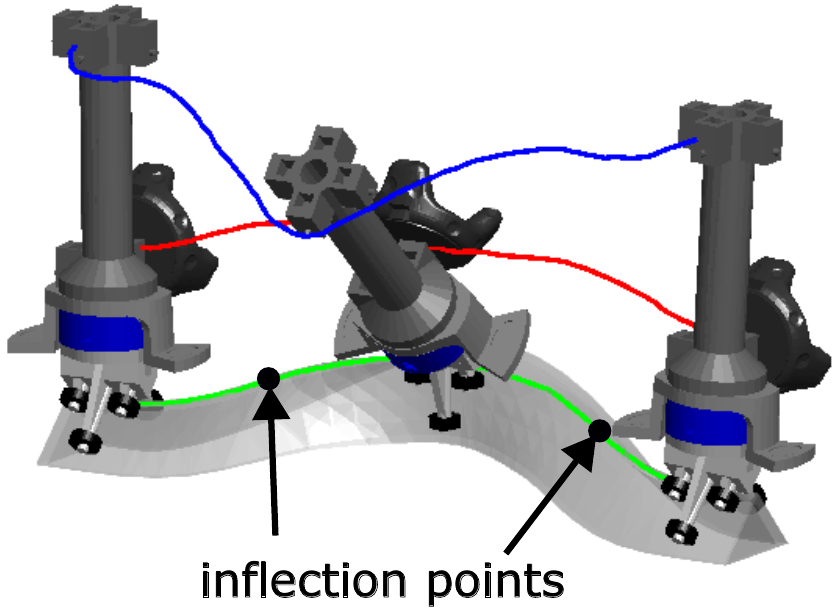}
		\caption{Real recording of a human-demonstrated contour-following task. The point trajectory of the tracker is shown in red, while the trajectories of two additional reference points on the object are shown in green and blue.}
		\vspace{-4pt}
		\label{fig:contour}
	\end{figure}

	This advantage is demonstrated by calculating $d_{k+1,k}$ for the rigid-body trajectory depicted in Figure~\ref{fig:contour}. 
	This trajectory corresponds to a real recording of a human-demonstrated contour-following task. The complete pose trajectory of the tool was captured using an HTC Vive Motion-Capture system, with a tracker attached to the tool. The point trajectory of the tracker is shown in red, while the trajectories of two additional reference points on the object are shown in green and blue. While following the contour, the motion ISA of the tool abruptly shifts in direction at two points along the contour~\cite{vochten2023invariant}. We refer to these points as inflection points.
	
	We calculated $d_{k+1,k}$ for these three pose trajectories with different choices for the body reference point. For this calculation, we applied a rule of thumb for $L$, setting it to approximately three times the largest distance between any two points on the tool ($\approx 90$cm). This ensured that regularization would only activate when the functional frame diverged a significant distance from the moving body, while still remaining within a sensible proximity. Additionally, we selected $\xi = 1\Delta s$, which resulted in a fine progress scale, particularly advantageous for localized, precise segmentation.
	
\begin{figure}[t]
		\centering
		\includegraphics[width = 0.65\linewidth]{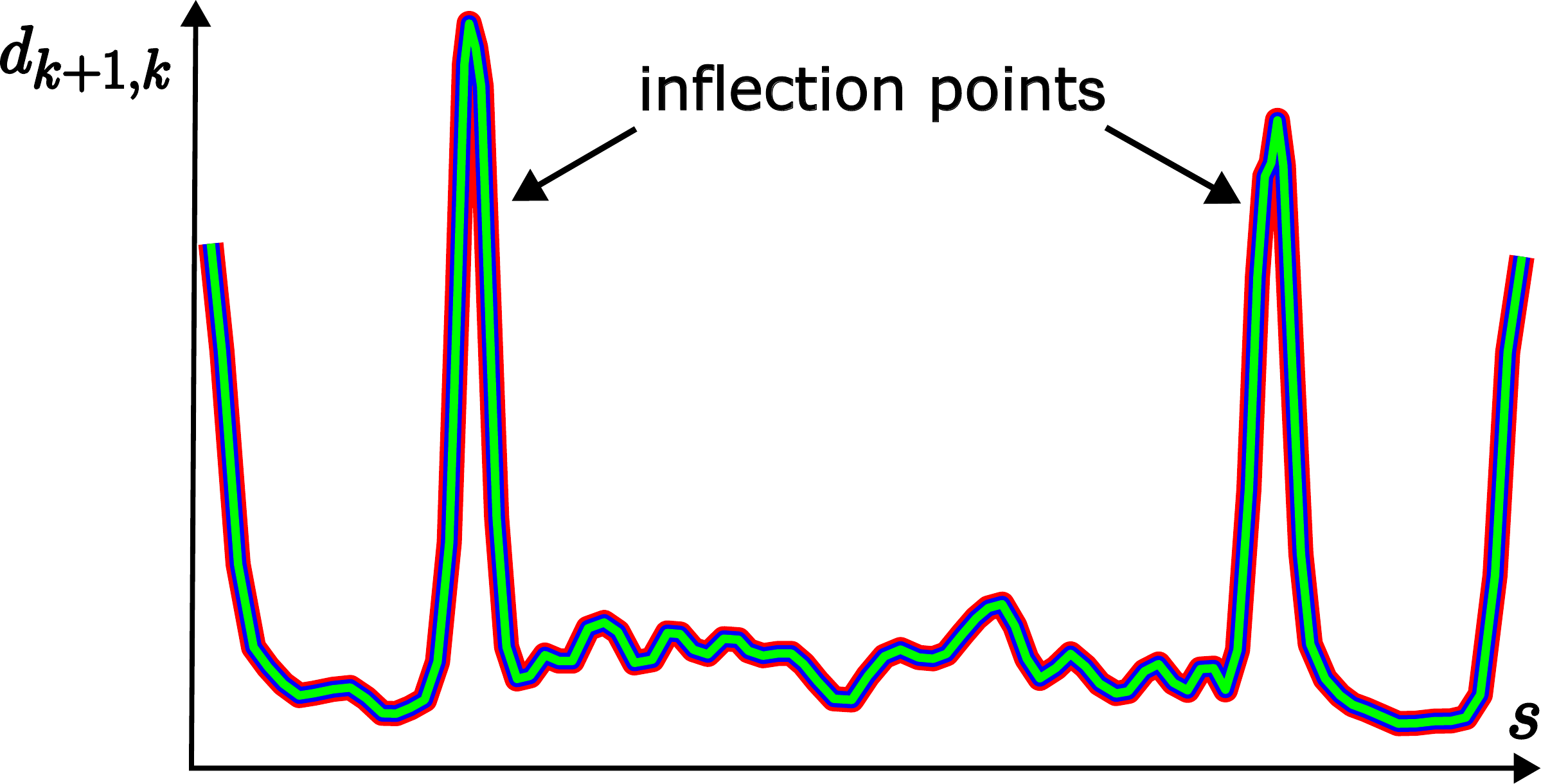}
		\caption{Visualisation of the three signals $d_{k+1,k}$ corresponding to the three pose trajectories with different choices for the body reference point as visualized in Figure~\ref{fig:contour}. }
		\vspace{-8pt}
		\label{fig:segmentation}
	\end{figure}

	Despite the variations in the point trajectories caused by changes in the body reference point on the tool, the computed signals $d_{k+1,k}$ remained identical, as shown in Figure~\ref{fig:segmentation}. 
	
	The result shows peaks in the change in trajectory shape at the inflection points along the contour. Such peaks can be detected and utilized for trajectory segmentation. In conclusion, these findings reconfirm that the BILTS measure is robust to variations in the body reference point. Furthermore, it demonstrates BILTS potential for the segmentation of trajectories, independent of reference frame selection and calibration.
	
	\vspace{3pt}
	\subsubsection{Application to trajectory adaptation}
	
	Besides trajectory recognition and segmentation, the BILTS measure can also be applied to trajectory adaptation towards novel contexts~\cite{vochten2019generalizing}. Within the framework of robot learning by demonstration, the objective is to adapt motions to novel situations while maintaining similarity with the original demonstration(s). Invariant similarity measures facilitate the robust evaluation of similarity between demonstrated and adapted trajectories, even under substantial reference frame variations. Hence, this robustness enables more effective extrapolation of demonstrated trajectories compared to traditional, non-invariant similarity measures. 
	
	In \cite{vochten2019generalizing}, demonstrated pose trajectories are adapted by minimizing the difference between the invariant descriptors (eFS or ISA) of the adapted and demonstrated trajectories. Alternatively, the proposed BILTS measure between the adapted and demonstrated trajectories could be minimized to achieve similar performance.
	An advantage of the BILTS measure for trajectory adaptation is that it requires the tuning of only two parameters ($L$ and $\xi$), as opposed to the six tuning parameters in \cite{vochten2019generalizing}. The reduced number of parameters, along with the physical interpretability of the parameters $L$ and $\xi$, offers potential to simplify the tuning process. 
	
	\vspace{3pt}
	\subsubsection{Application to force trajectories}
	
	The focus of this paper was on measuring the similarity between \textit{motion} trajectories. Nevertheless, the BILTS measure is also directly applicable to \textit{force} trajectories (e.g. screw wrenches) due to the duality between motion and force~\cite{vochten2023invariant}. Therefore, the BILTS measure can also be used to measure the similarity between force trajectories. This extension could be valuable for segmenting trajectories based on varying contact geometries and recognizing patterns in contact forces. Such capabilities could enhance the analysis of interactions involving contact forces.

	\subsection{Conclusion}
	We introduced the novel continous-time and discretized \textit{Bi-Invariant Local Trajectory-Shape Similarity} (BILTS) measures for rigid-body motion, which quantify the difference in trajectory shape as a single scalar value ($d$ or $d_k$) in a bi-invariant manner. Compared to existing measures, the BILTS measure is the only one that posesses both bi-invariance and boundedness, and that captures third-order shape identity. The discretized BILTS measure $d_k$ can be directly computed from discrete-time rigid-body trajectories, without requiring explicit estimation of higher-order trajectory derivatives. Through rigorous recognition experiments using multiple datasets, we demonstrated the regularized BILTS$^+$ measure's exceptional robustness to singularities. On average, BILTS$^+$ also attained the highest recognition ratio of 95.3\% compared to other State-of-the-Art similarity measures (DHB, eFS, ISA, ISA-ocp, RRV, DSRF). It also exhibited the least sensitivity to contextual variations, as indicated by its low standard deviation of 3.4\% across the various datasets. We believe that the BILTS measure is a valuable tool for recognizing motions performed in diverse contexts and has potential applications in other trajectory analysis tasks, including the recognition, segmentation, and adaptation of both motion and force trajectories.

	\section*{Acknowledgements}
	This result is part of a project that has received funding from the European Research Council (ERC) under the European Union's Horizon 2020 research and innovation programme (Grant agreement No. 788298).

	\normalfont
	\bibliographystyle{IEEEtran}
	\bibliography{ref_v2}

\end{document}